\documentclass[times,twocolumn,final]{elsarticle}

\usepackage{cite}
\usepackage{amsmath,amssymb,amsfonts}
\usepackage{algorithmic}
\usepackage{graphicx}
\usepackage{textcomp}

\usepackage{medima}
\usepackage{framed,multirow}

\usepackage{amssymb}
\usepackage{latexsym}

\usepackage{url}
\usepackage{xcolor}

\usepackage{hyperref}

\definecolor{newcolor}{rgb}{.8,.349,.1}

\journal{Medical Image Analysis}

\begin{document}

\verso{Given-name Surname \textit{et~al.}}

\begin{frontmatter}

\title{Marginal loss and exclusion loss for partially supervised multi-organ segmentation}

\author[1,2]{Gonglei \snm{Shi}}

\author[1]{Li \snm{Xiao}\corref{cor1}}
\ead{xiaoli@ict.ac.cn}

\author[2]{Yang \snm{Chen}}

\author[1]{S. Kevin \snm{Zhou}\corref{cor1}}

\cortext[cor1]{Corresponding author:}
\ead{zhoushaohua@ict.ac.cn}

\address[1]{Medical Imaging, Robotics, Analytic Computing Laboratory \& Engineering (MIRACLE) Group, \\ Institute of computing technology, Chinese Academy of Sciences, Beijing, 100190,China}
\address[2]{School of Computer Science and Engineering, Southeast University, Nanjing, 210000, China}


\begin{abstract}
Annotating multiple organs in medical images is both costly and time-consuming; therefore, existing multi-organ datasets with labels are often low in sample size and mostly partially labeled, that is, a dataset has a few organs labeled but not all organs. In this paper, we investigate how to learn a single multi-organ segmentation network from a union of such datasets. To this end, we propose two types of novel loss function, particularly designed for this scenario: (i) marginal loss and (ii) exclusion loss. Because the background label for a partially labeled image is, in fact, a `merged' label of all unlabelled organs and `true' background (in the sense of full labels), the probability of this `merged' background label is a marginal probability, summing the relevant probabilities before merging. This marginal probability can be plugged into any existing loss function (such as cross entropy loss, Dice loss, etc.) to form a marginal loss. Leveraging the fact that the organs are non-overlapping, we propose the exclusion loss to gauge the dissimilarity between labeled organs and the estimated segmentation of unlabelled organs. Experiments on a union of five benchmark datasets in multi-organ segmentation of liver, spleen, left and right kidneys, and pancreas demonstrate that using our newly proposed loss functions brings a conspicuous performance improvement for state-of-the-art methods without introducing any extra computation.
\end{abstract}

\begin{keyword}
\KWD Multi-organ segmentation\sep partially labeled dataset\sep marginal loss\sep exclusion Loss
\end{keyword}

\end{frontmatter}


\section{Introduction}
\label{sec:introduction}
Multiple organ segmentation has been widely used in clinical practice, including diagnostic interventions, treatment planning, and treatment delivery~ \citep{Ginneken2011Computer,Sykes2014Reflections}. It is a time-consuming task in radiotherapy treatment planning, with manual or semi-automated tools~\citep{Heimann2009Comparison} frequently employed to delineate organs at risk.  Therefore, to increase the efficiency of organ segmentation, auto-segmentation methods such as statistical models~\citep{2015Automatic,okada2015abdominal}, multi-atlas label fusion~\citep{xu2015efficient,tong2015discriminative,suzuki2012multi}, and registration-free methods~\citep{Saxena2016Automated,Lombaert2014Laplacian,Baochun2015fully} have been developed. Unfortunately, these methods are likely affected by image deformation and inter-subject variability and their success in clinical applications is limited.

Deep learning based medical image segmentation methods have been widely used in the literature to perform the classification of each pixel/voxel for a given 2D/3D medical image and has significantly improved the performance of multi-organ auto-segmentation.
One prominent model is U-Net~\citep{ronneberger2015u}, along with its latest variant nnUNet~\citep{isensee2018nnu}, which learns multiscale features with skip connections. Other frameworks for multi-organ segmentation include~ \citep{wang2019abdominal,binder2019multi,gibson2018automatic}. There is a rich body of subsequent works~ \citep{okada2012multi,chu2013multi,suzuki2012multi,liu2020deep,gibson2018automatic}, focusing on improving existing frameworks by finding and representing the interrelations based on canonical correlation analysis especially by constructing and utilizing the statistical atlas. 

\begin{figure}[]
  \begin{center}
    \includegraphics[width=\linewidth]{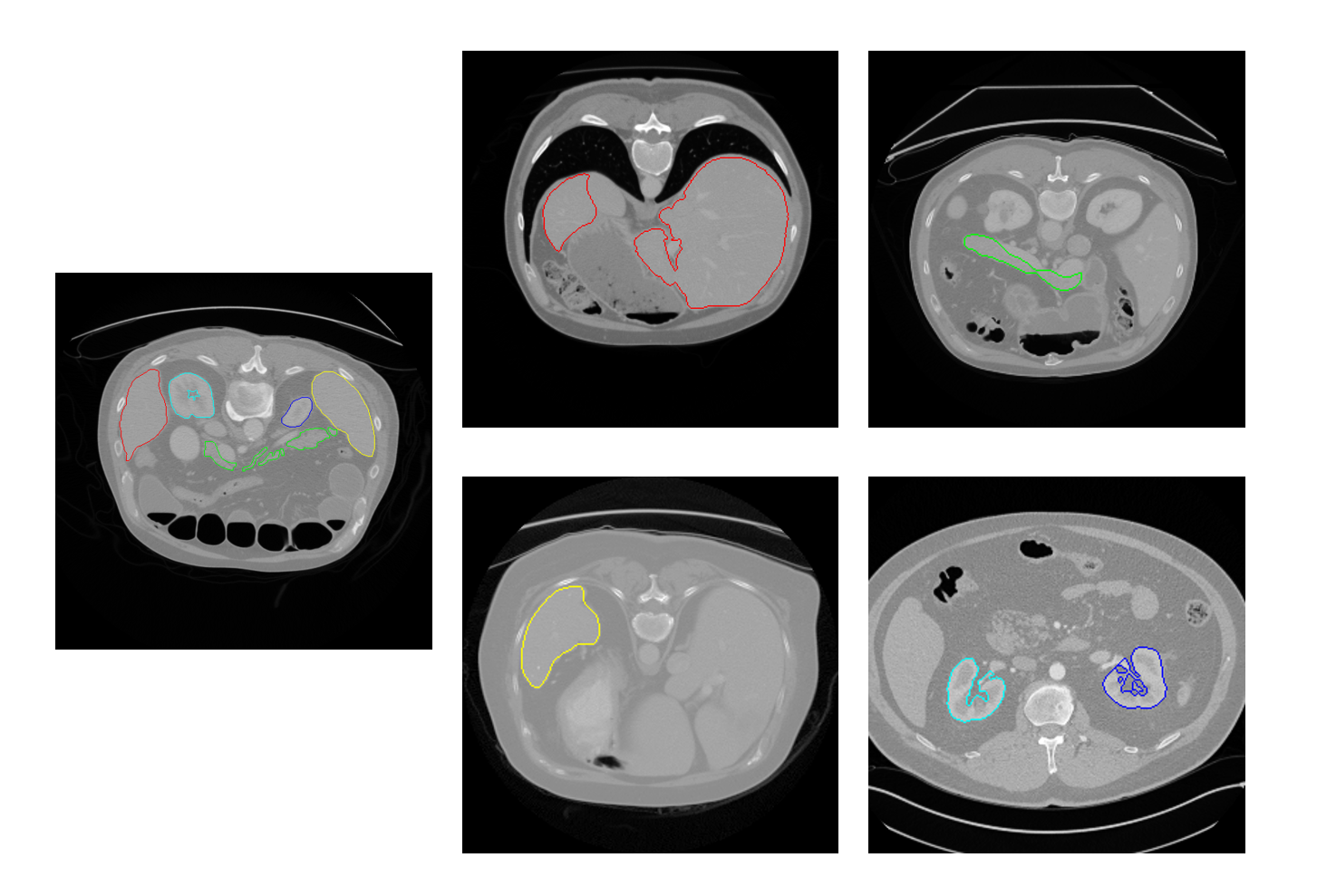}
    \caption{Five typical annotated images from five different datasets, one image per dataset. The colored edges show the annotated organ boundaries (red for liver, yellow for spleen, green for pancreas, blue for left kidney, and cyan for right kidney). The left image shows the case of a fully annotated data set and the amount of such data set is usually small. The right four images are partially labeled.}
  \end{center}
  \label{datascheme1}
\end{figure}

However, almost all current segmentation models rely on fully annotated data~ \citep{zhao2019data,chen2018voxresnet,Liverseg} with strong supervision. To curate a large-scale fully annotated dataset is a challenging task, both costly and time-consuming. It is also a bottleneck in the multi-organ segmentation research area that current labeled data sets are often low in sample size and mostly partially labeled. That is, a data set has a few organs labeled but not all organs (as shown in Fig.~\ref{datascheme1}).  Such partially annotated datasets obviate the use of segmentation methods that require full supervision.

It becomes a research problem of practical need on how to make full use of these partially annotated data to improve the segmentation accuracy and robustness. In the case of sufficient network model capabilities, a larger amount of data typically means that it is more likely to represent the actual distribution of data in reality, hence leading to better overall performance. Motivated by this, in this paper we investigate how to learn a \textbf{single multi-organ segmentation network} from the union of such partially labeled data sets. Such learning does not introduce any extra computation.

To this end, we propose two types of loss functions particularly designed for this task: (i) \textbf{marginal loss} and (ii) \textbf{exclusion loss}. Firstly, because the background label for a partially labeled image is, in fact, a `merged' label of all unlabeled organs and `true' background (in the sense of full labels), the probability of this `merged' background label is a marginal probability, summing the relevant probabilities before merging. This marginal probability can be plugged into any existing loss function such as cross entropy (CE) loss, Dice loss, etc. to form a \textbf{marginal loss}. In this paper, we propose to use marginal cross entropy loss and marginal Dice loss in the experiment. Secondly, in multi-organ segmentation, there is a one-to-one mapping between pixels and labels, different organs are mutually exclusive and not allowed to overlap. This leads us to propose the \textbf{exclusion loss}, which adds the exclusiveness as prior knowledge on each labeled image pixel. In this way, we make use of the explicit relationships of given ground truth in partially labeled data, while mitigating the impact of unlabeled categories on model learning. Using the state-of-the-art network model (e.g., nnUNet~\citep{isensee2018nnu}) as the backbone, we successfully learn a single multi-organ segmentation network that outputs the full set of organ labels (plus background) from a union of five benchmark organ segmentation datasets from different sources. Refer to Fig.~\ref{datascheme1} for image samples from these datasets.

In the following, after a brief survey of related literature in Section~\ref{sec.paper}, we provide the derivation of marginal loss and exclusion loss in Section~\ref{sec.method}. The two types of loss function can be applied to pretty much any loss function that relies on posterior class probabilities. In Section~\ref{sec.exp}, extensive experiments are then presented to demonstrate the effectiveness of the two loss functions. By successfully pooling together partially labeled datasets, our new method can achieve significant performance improvement, which is essentially \textbf{a free boost} as these auxiliary datasets are existent and already labeled. Our method outperforms two state-of-the-art models~\citep{zhou2019prior,tmipld} for partially annotated data learning. We conclude the paper in Section~\ref{sec.conc}.

\section{Related Work} \label{sec.paper}
\subsection{Multi-organ segmentation models}
Many pioneering works have been done on multi-organ segmentation, using traditional machine learning methods or deep learning methods. In~\citep{okada2015abdominal,xu2015efficient,tong2015discriminative,suzuki2012multi,shimizu2007segmentation,altas1}, a multi-altas based strategy is used for segmentation, which registers an unseen test image with multiple training images and use the registration map to propagate the labels in the training images to generate final segmentation. However, its performance is limited by image registration quality. In~\citep{heimann2009statistical,statistical2,statistical3}, prior knowledge of statistical models is employed to achieve multi-organ segmentation. There are also some methods that directly use deep learning semantic segmentation networks for multi-organ segmentation~\citep{gibson2018automatic,wang2019abdominal,multi_levelset,Contextmodel}. Besides, 
there are prior approaches that combine the above-mentioned different methods~\citep{chu2013multi,other1} to achieve better multi-organ segmentation. However, all these methods rely on the availability of fully labelled images.

\subsection{Multi-organ segmentation with partially annotated data learning}
Very limited works have been done on medical image segmentation with partially-supervised learning. Zhou et al.~\citep{zhou2019prior} learns a segmentation model in the case of partial labeling by adding a prior-aware loss in the learning objective to match the distribution between the unlabeled and labeled datasets. However, it trains separate models for the fully labeled and partially labeled datasets, and hence involves extra memory and time consumption. 
Instead, our work trains a single multi-class network. Since only two loss terms are added, it needs nearly no additional training time and memory cost. Dmitriev et al.~ \citep{Dmitriev_2019_CVPR} propose a unified, highly efficient segmentation framework for robust simultaneous learning of multi-class datasets with missing labels. But the network can only learn from datasets with single-class labels. Fang et al.~ \citep{tmipld} hierarchically incorporate multi-scale features at various depths for image segmentation, further develop a unified segmentation strategy to train three separate datasets together, and finally achieve multi-organ segmentation by learning from the union of partially labeled and fully labeled datasets. Though this paper also uses a loss function that amounts to our marginal cross entropy, its main focus is on proposing the hierarchical network architecture. In contrast, we concentrate on studying the impact of the marginal loss including both marginal cross entropy and marginal Dice loss. Furthermore, it is worth mentioning that none of the above works considers the mutual exclusiveness, a well-known attribute between different organs. We propose a novel exclusion loss term, exploiting the fact that organs are mutually exclusive and adding the exclusiveness as prior knowledge on each image pixel. 

\subsection{Partially annotated data learning in other tasks}
A few existing methods have been developed on classification and object detection tasks using partially annotated data.  Yu et al.~\citep{yu2014large} propose an empirical risk minimization framework to solve multi-label classification problem with missing labels; Wu et al.\citep{wu2015multi} train a classifier with multi-label learning with missing labels to improve object detection problem.  Cour et al.~\citep{JMLR:v12:cour11a} propose a convex learning formulation based on the minimization of a loss function appropriate for the partially labeled setting.  
Besides, as far as semi-supervised learning is concerned, a number of researches have been developed to solve~ \citep{he2019joint,zhu2018multi,Li2019} classification problems or detection problems in the absence of annotations.

\section{Method} \label{sec.method}

The goal of our work is to train a single multi-class segmentation network $\Psi$ by using a large number of partially annotated data in addition to a few fully labeled data for baseline training. Learning under such a setup is enabled by the novel losses we propose below. 

Segmentation is achieved by grouping pixels (or voxels) of the same label. A labeled pixel has two attributes: (i) pixel and (ii) label.
Therefore, it is possible to improve the segmentation performances by exploiting the pixel or label information. To be more specific, we leverage some prior knowledge on each image pixel, such as its anatomical location or its relation with other pixels, to facilitate the network for better segmentation; we also merge or split labels to help the network focus more on specific task requirements. In this work, we apply the two ideas on multi-organ segmentation tasks as follows. Firstly, due to a large amount of partially labeled images, we merge all unlabeled organ pixels with the background label, which forms a \textbf{marginal loss}. Secondly, regarding a well known prior knowledge that organs are mutually exclusive, we design an \textbf{exclusion loss}, which adds exclusion information on each image pixel, to further reduce the segmentation errors.

\subsection{Regular cross-entropy loss and regular Dice loss}
The loss function is generally proposed for a specific problem. A common idea for loss functions are based on classification tasks which optimize the intra-class difference and reduce the intra-class variation, for example contrastive loss~ \citep{hadsell2006dimensionality}, triplet Loss~
\citep{schroff2015facenet}, center loss~\citep{wen2016discriminative}, large margin softmax loss~\citep{liu2016large}, angular softmax~\citep{li2018angular} and cosine embeding loss~\citep{wang2018cosface}. The cross entropy loss~\citep{Long_2015_CVPR} is the most representative loss function, which is commonly used in deep learning. There are also some loss functions designed to optimize the global performance for semantic segmentation, such as Dice loss~ \citep{Long_2015_CVPR}, Tversky loss~
\citep{salehi2017tversky}, combo loss~\citep{taghanaki2019combo}, Lovasz-Softmax loss~\citep{berman2018lovasz}. Besides, some losses are proposed specifically to improve a given loss function, for example, the focal loss~\citep{lin2017focal} is developed based on cross-entropy loss~\citep{Long_2015_CVPR} to better solve class imbalance problem. Here we focus on the cross-entropy loss and regular Dice loss that are most commonly used in multi-organ segmentation.

Suppose that, for a multi-class classification task with $N$ labels with its label index set as $\Omega_N=\{C_1,C_2,\ldots,C_N\}$, its data sample $x$ (i.e., an image pixel in image segmentation) belongs to one of $N$ classes, say class $C_n$, which is encoded as an $N$-dimensional one-hot vector ${\hat y}_n=[y_1,y_2,\ldots,y_N]$ with $y_n=1$ and all others $0$. A multi-class classifier consists of a set of response functions $\{a_n(x);~n\in \Omega_N\}$, which constitutes the outputs of the segmentation network $\Psi$. 
From these response functions, the posterior classification probabilities are computed by a softmax function,
\begin{equation}
    p_n = \frac {\exp(a_n)} {\sum_{k \in \Omega_N} \exp(a_k)}, ~~ n \in\Omega_N .
    \label{softmax}
\end{equation}

To learn the classifier, the regular cross-entropy loss  is often used, which is defined as follows: 
\begin{equation}
    \label{r2}
    L_{rCE} = - \sum_{n \in \Omega_N} y_n \log(p_n).
\end{equation}

Besides, the Dice score coefficient (DSC) is often used, which measures the overlap between the segmentation map and ground truth. The dice loss is defined as $1-DSC$:
\begin{equation}
    \label{r3}
    L_{rDice} = \sum_{n\in\Omega_N}(1-2\cdot\frac{y_n~ p_n}{y_n+p_n})
\end{equation}

\subsection{Marginal loss}

For an image with incomplete segmentation label, it is possible that the pixels for some given classes are not `properly' provided. To deal with such situations, we assume that there are a reduced number of $M<N$ classes in a partially-labeled dataset with its corresponding label index set as $\Omega'_M=\{C'_1,C'_2,\ldots,C'_{M}\}$. For each merged class label $m \in \Omega'_{M}$, there is a corresponding subset ${\Phi_{m}} \subset \Omega_N$, which is comprised of all the label indexes in $\Omega_{N}$ that can be merged into the same class $m$. Because the labels are exclusive in multi-organ segmentation, we have $\Omega_N = \cup_{m \in \Omega'_{M}} {\Phi_{m}}$.

\begin{figure}[h]
    \centering
    \includegraphics[width=\linewidth]{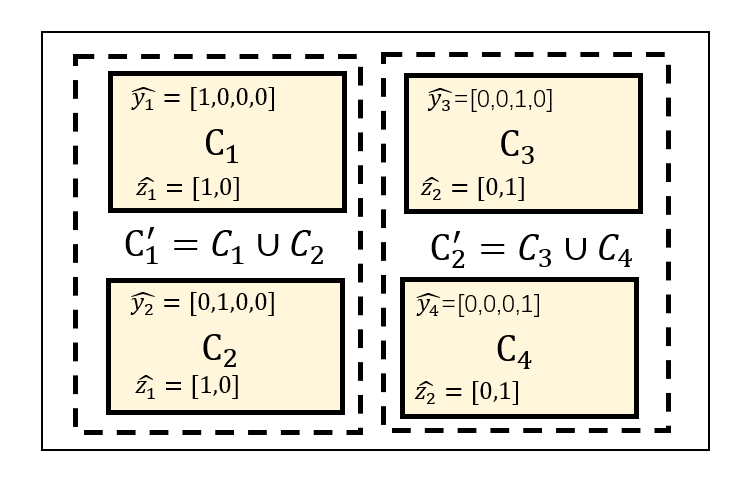}
    \caption{Venn Diagram to illustrate the marginal loss. The dataset contains three classes $C_1,C_2,C_3,C_4$, the partially labeled dataset only contains two class labels, with $C_1$ and $C_2$ merged together as $C'_1$ and $C_3$ and $C_4$ merged together as $C'_2$.}
    \label{mVenn1}
\end{figure}

Fig.\ref{mVenn1} illustrates the process of label merging, using an example of four organ classes  $C_i, i=1,2,3,4$. 
After the merging, there are two classes $C'_1$ and $C'_2$, with $C_1$ and $C_2$ are combined together to form a new merged label $C'_1$ and $C_3$ and $C_4$ to form a new label $C'_2$.

The classification probability for the merged class $m$ is a marginal probability
\begin{equation}
    q_{m} = \sum_{n \in {\Phi_{m}}}  p_n.
    \label{marginalmerge}
\end{equation}
Also, the one-hot vector for a class $m \in \Omega'_{M}$ is denoted as ${\hat z}_m=[z_1,z_2,\ldots,z_M]$, which is $M$-dimensional with $z_m=1$ and all others $0$.

Consequently, we define marginal cross-entropy loss and marginal Dice loss as follows:
\begin{equation}
    \label{m2}
    L_{mCE} = - \sum_{m \in \Omega'_{M}} z_{m} \log(q_{m}).
\end{equation}
\begin{equation}
    \label{m3}
    L_{mDice} = \sum_{m\in \Omega'_{M}}(1-2\cdot\frac{z_m ~   q_m}{z_m+q_m}).
\end{equation}

We use marginal cross entropy as an example to perform the gradient calculation. Firstly, referring to Eqs. (\ref{softmax}) and (\ref{marginalmerge}), the gradient of the output $m$ of a softmax node to the network node $a_j$ is:
\begin{equation}
        \frac{\partial q_{m}}{\partial a_j} = \sum_{n\in \Phi_{m}}{\frac{\partial p_n}{\partial a_j}} 
        = p_j [\pi_j(\Phi_{m}) -q_m ],
\end{equation}
where $\pi_j(\Phi_{m})$ is a boolean indicator function that tells if $j$ is in $\Phi_{m}$. $p$ and $q$ are the classification probabilities of regular and marginal softmax functions.The derivative gradient of $L_{mCE}$ to the network node $a_{j}$ is:
\begin{equation}
    \begin{aligned}
        & \frac{\partial L_{mCE}}{\partial a_j} 
        =   -\sum_{m\in \Omega'_{M}} \frac{z_{m}}{q_{m}}  \frac{\partial q_{m}}{\partial a_j} \\
        = &  - \sum_{m\in \Omega'_{M}} \frac{z_{m}}{q_{m}} p_j [\pi_j(\Phi_{m}) -q_m ] 
        = [1-\frac{z_{\bar m}}{q_{\bar m}}]~p_j,\\
    \end{aligned}
\end{equation}
where ${\bar m}$ is the only class index that makes $\pi_j(\Phi_{m})=0$.

\subsection{Exclusion loss}
\begin{figure}[h]
    \centering
    \includegraphics[width=\linewidth]{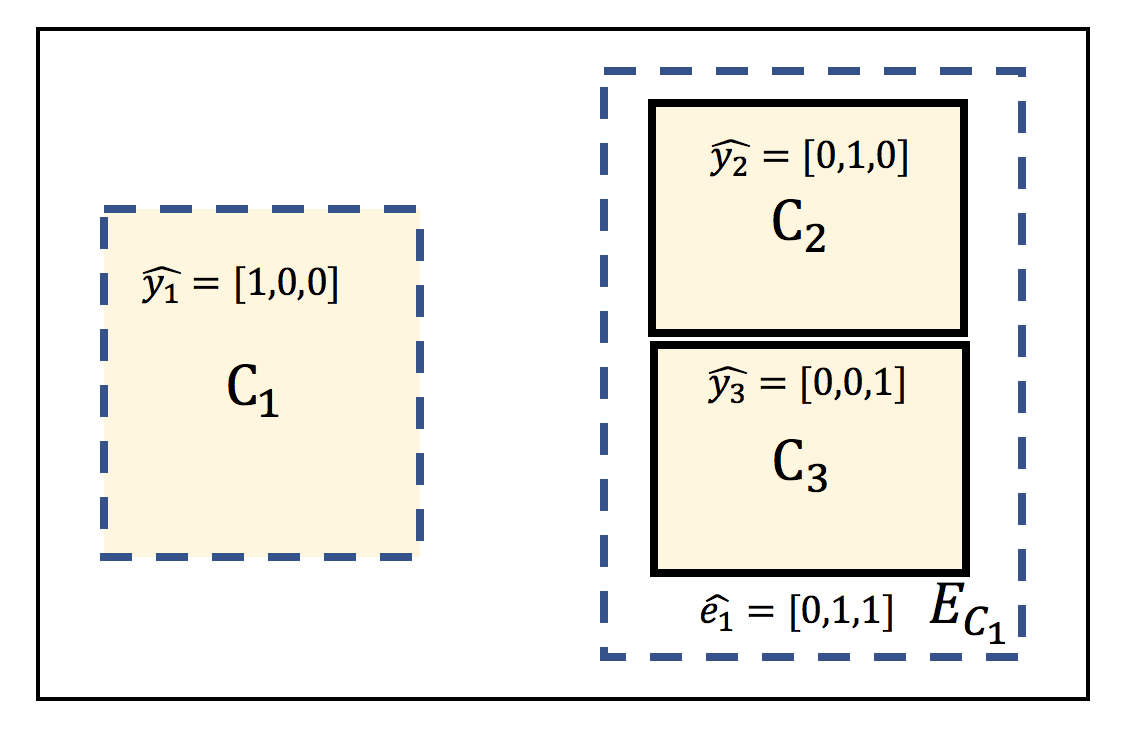}
    \caption{Venn Diagram to illustrate the exclusion loss. There are three mutually exclusive classes $C_1$, $C_2$, and $C_3$. The exclusion set for $C_1$ is  $E_{C_1}=C_2\cup C_3$.} 
    \label{eVenn2}
\end{figure}

It happens in multi-organ segmentation tasks that some classes are mutually exclusive to each other. The exclusion loss is designed to add the exclusiveness as an additional prior knowledge on each image pixel. 
We define an exclusion subset for a class $n$ as $E_n$, which comprises all (or a part of) the label indexes that are mutually exclusive with class $n$. The exclusion label information is encoded as an N-dimensional vector ${\hat e}_n=[e_1,e_2,\ldots,e_N]$, which is obtained as: 
\begin{equation}
        {\hat e}_n = \sum_{k\in E_{n}} {\hat y}_k. 
\end{equation}
Note that ${\hat e}_n$ is still an $N$-dimensional vector, but it is not an one-hot vector any more. Fig.\ref{eVenn2} shows the procedure of applying exclusion loss. Assuming that organ classes $C_1$, $C_2$ and $C_3$ are mutually exclusive, the labels of $C_2$ and $C_3$ form the exclusion subset $E_{C_1}$. 

We expect that the intersection between the segmentation prediction $p_n$ from the network and $e_n$ is as small as possible. Following the Dice coefficient, the formula for the exclusion Dice loss is given as:
\begin{equation}
    \label{e}
    L_{eDice}=\sum_{{n}\in\Omega_N}2\cdot\frac{e_n \cdot p_n}{e_n+p_n}.
\end{equation}
The exclusion cross-entropy loss is defined accordingly:
\begin{equation}
\label{eSoft}
L_{eCE}=\sum_{n\in\Omega_N} e_n \log(p_n+\epsilon),
\end{equation}
where $\epsilon$ is introduced to avoid the trap of $-\infty$. We set $\epsilon=1$.

\begin{figure*}[t]
    \centering
    \includegraphics[width=0.85\textwidth]{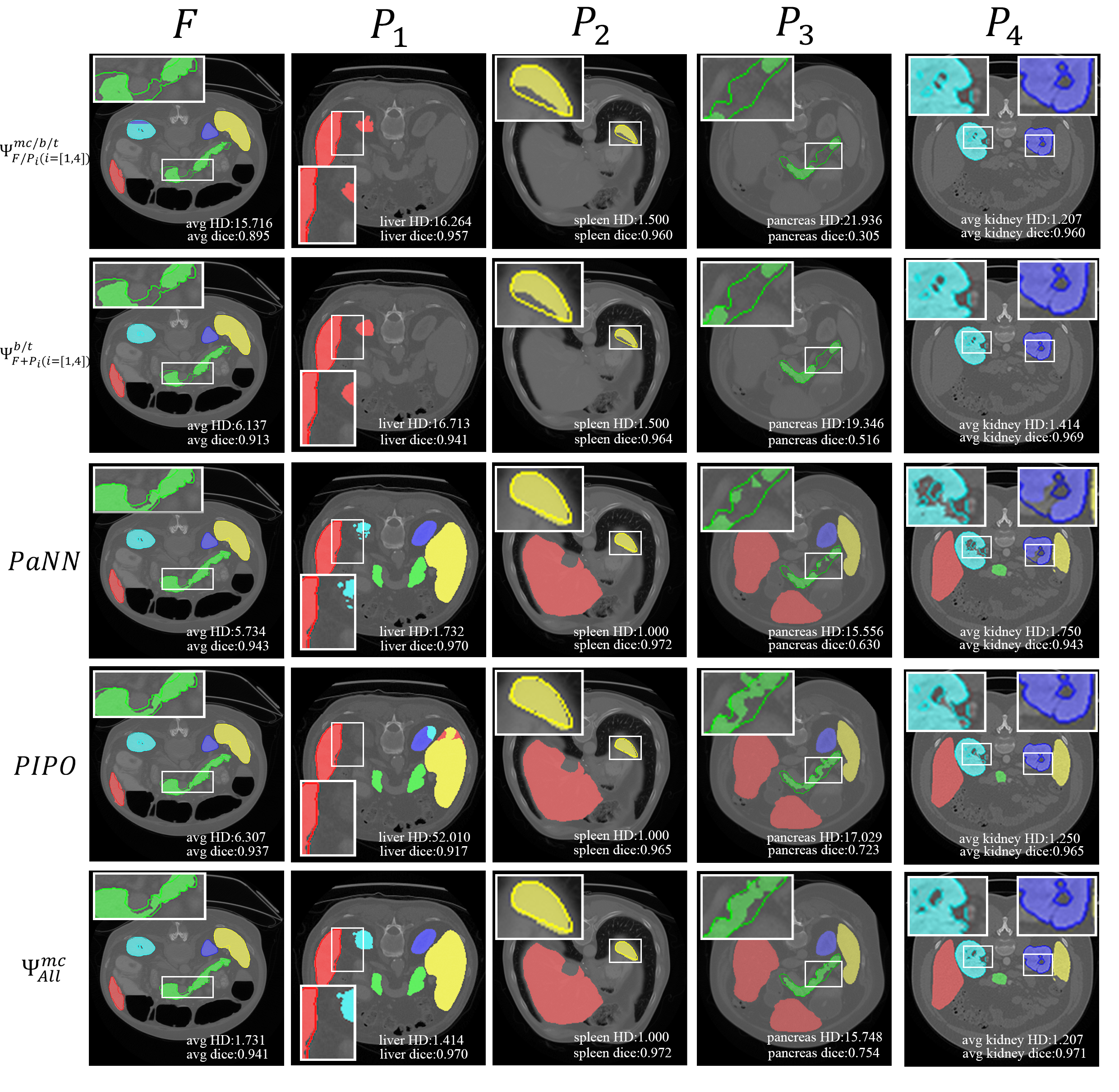}
    \caption{The comparison of results obtained by different segmentation networks. 
    The red area represents the liver, the yellow area represents the spleen, the green area represents the pancreas, the cyan and blue areas represent the right and left kidneys, respectively. The edge with deeper color means the ground truth given by the dataset.}
    \label{present}
\end{figure*}

\begin{table*}[t]
    \centering
    \caption{Usage of experimental dataset}
    \resizebox{\linewidth}{!}{
    \begin{tabular}{l|cccccccccc}
    \hline
    Network & Liver$\in{F}$ & Liver$\in{P_1}$ & Spleen$\in{F}$ & Spleen$\in{P_2}$ &Pancreas$\in{F}$&Pancreas$\in{P_3}$& L Kidney$\in{F}$& R Kidney$\in{F}$& L Kidney$\in{P_4}$ & R Kidney$\in{P_4}$ \\ \hline
   $\Psi^{mc}_F$: multiclass ($F$) & $\surd$       &            & $\surd$      & & $\surd$      &  & $\surd$       &$\surd$            &       &       \\ \hline
   $\Psi^{b}_{F+P_1}$: binary liver ($F+P_1$) & $\surd$ & $\surd$&&&&&&&& \\ \hline
    $\Psi^{b}_{F+P_2}$: binary spleen ($F+P_2$)&&& $\surd$  & $\surd$ &&&&&&    \\ \hline
    $\Psi^{b}_{F+P_3}$: binary pancreas ($F+P_3$)&&&&& $\surd$  & $\surd$ &&&&    \\ \hline
    $\Psi^{b}_{F+P_4}$: binary kidney ($F+P_4$)&&&&&&& $\surd$  & $\surd$ & $\surd$  & $\surd$    \\ \hline
    $\Psi^b_{P_1}$: binary liver ($P_1$) && $\surd$ &&&&&&&&             \\ \hline
    $\Psi^b_{P_2}$: binary spleen ($P_2$) &&&& $\surd$ &&&&&&    \\ \hline
    $\Psi^{b}_{P_3}$: binary pancreas ($P_3$)&&&&&& $\surd$ &&&&    \\ \hline
    $\Psi^{t}_{P_4}$: ternary kidney ($P_4$)&&&&&&&&& $\surd$  & $\surd$    \\ \hline
    $\Psi^{mc}_{All}$: multiclass ($F+P_{1:4}$)       & $\surd$       & $\surd$          & $\surd$        & $\surd$  & $\surd$       & $\surd$& $\surd$       & $\surd$& $\surd$       & $\surd$         \\ \hline
    total \# of training CT      &24         &100           &24          &33&24  &224&24&24&168&168 \\ \hline
    total \# of testing CT      &6         &26           &6          &8&6  &56 &6&6&42&42 \\ \hline
    \end{tabular}
    }
    \label{usage}
\end{table*}

\section{Experiments and Results} \label{sec.exp}
\subsection{Problem setting and benchmark dataset}
We consider a partially-supervised multi-organ segmentation task that is common in practice (such as Fig.~\ref{datascheme1}). For each partially annotated image, we restrict it with only one label. For clarity of description, we assume that $F$ denotes the fully-labeled segmentation dataset and $P_i; i\in\{1,2,\ldots,C\}$ denotes a dataset of partially-annotated images that contain only a partial list of organ label(s). The datasets $P_{1:C}$ do not overlap in terms of their organ labels. For an image in $P_i$, there is a `merged' background, which is the union of real background and missing organ labels. We jointly learn a single segmentation network $\Psi$ using $F \cup P_1 \cup ...\cup P_C$, assisted by the proposed loss functions.

For our experiments, we choose liver, spleen, pancreas, left kidney and right kidney as the segmentation targets and use the following benchmark datasets. 
\begin{itemize}
    \item{Dataset $F$}. We use Multi-Atlas Labeling Beyond the Cranial Vault - Workshop and Challenge~\citep{landman2017multi} as fully annotated base dataset $F$. It is composed of 30 CT images with segmentation labels of 13 organs, including liver, spleen, right kidney, left kidney, pancreas, and other organs (gallbladder, esophagus, stomach, aorta, inferior vena cava, portal vein and splenic vein,  right adrenal gland, and left adrenal gland) we hereby ignore.  
    \item{Dataset $P_1$}. We refer to the task03 liver dataset from the Decathlon-10 \citep{simpson2019large} challenge as $P_1$. It is composed of 130 CT's with annotations for liver and liver cancer region. We merge the cancer label into the liver label and obtain a binary-class (liver vs background) dataset.
    \item{Dataset $P_2$}. We refer to the task09 spleen dataset from the Decathlon-10 challenge as $P_2$. It includes 41 CT's with spleen segmentation label.
    \item{Dataset $P_3$}. We refer to the task07 pancreas dataset from the Decathlon-10 challenge as $P_3$. It includes 281 CT's with pancreas and its cancer segmentation label. The cancer label is merged into the pancreas label to obtain a binary-class (pancreas vs background) dataset.
    \item{Dataset $P_4$}. We refer to KiTS~\citep{KiTS} challenge dataset as $P_4$. Since the offered 210 CT segmentation makes no distinction between left and right kidneys, we manually divide it into left and right kidneys according to the connected component. Cancer label is merged into the according kidney label.
\end{itemize}
The spatial resolution of all these datasets are resampled to $(1.5\times 1.5\times 3)mm^3$. We split the datesets into training and testing. we randomly choose 6 samples from $F$, 26 samples from $P_1$ and 8 samples from $P_2$, 56 samples from $P_3$ and 42 samples from $P_4$ as testing. The others are used for training. Table \ref{tbl:data} also provides a summary description of the datasets.

\begin{table*}
    \caption{A summary description of the datasets.}
    \resizebox{\linewidth}{!}{
        \begin{tabular}{c|lllllllll|c} \hline
            Dataset & Modiality & Num of labeled samples & Annotated organs & axis & image voxel range&spacing range \\ \hline
             
             &&& liver / right kidney / left kidney / & z & $85\sim198$ & $2.50\sim5.00$ \\ 
             MALBCVWC& CT& 30 & /pancreas /spleen / other structures & y & 512 & $0.59\sim0.98$ \\
             &&&  & x & 512 & $0.59\sim0.98$ \\ \hline
             
             &&&&z &$ 74\sim984$ & $0.70\sim5.00$\\
             Decathlon-Liver& CT &126 &  liver & y & 512 & $0.56\sim1.00$ \\
             &&&&x&512&$0.56\sim1.00$\\ \hline
             
             &&&&z &$ 31\sim168$ & $1.50\sim8.00$\\
             Decathlon-Spleen& CT &41 &  spleen & y & 512 & $0.61\sim0.98$ \\
             &&&&x&512&$0.61\sim0.98$ \\ \hline
             
             &&&&z &$ 31\sim751$ & $0.70\sim7.50$\\
             Decathlon-Pancreas& CT &281 &  pancreas & y & 512 & $0.61\sim0.98$ \\
             &&&&x&512&$0.61\sim0.98$ \\ \hline
             
             &&&&z &$ 29\sim1059$ & $0.50\sim5.00$\\
             KiTS& CT &210 &  left kidney and right kidney  & y & 512 & $0.44\sim1.04$ \\
             &&&&x&512&$0.44\sim1.04$ \\ \hline
        \end{tabular}
    } \label{tbl:data}
\end{table*}

\subsection{Segmentation networks}

We set up the training of 10 deep segmentation networks for comparison as in Table~\ref{usage}. 
\begin{itemize}
    \item $\Psi^{mc}_F$: a multiclass segmentation network based on $F$.
    \item $\Psi^{b}_{P_1}$: a binary segmentation network for liver only based on $P_1$. 
    \item $\Psi^{b}_{P_2}$: a binary segmentation network for spleen only based on $P_2$.
    \item $\Psi^{b}_{P_3}$: a binary segmentation network for pancreas only based on $P_3$.
    \item $\Psi^{t}_{P_4}$: a ternary segmentation network for left kidney and right kidney only based on $P_4$.     
    \item $\Psi^{b}_{F+P_1}$: a binary segmentation network for liver only based on $F$ and $P_1$. Note that the spleen, pancreas, left kidney and right kidney labels in $F$ are merged into background.
    \item $\Psi^{b}_{F+P_2}$: a binary segmentation network for spleen only based on $F$ and $P_2$. Note that the liver, pancreas, left kidney and right kidney labels in $F$ are merged into background.
    \item $\Psi^{b}_{F+P_3}$: a binary segmentation network for pancreas only based on $F$ and $P_3$. Note that the liver, spleen, left kidney and right kidney labels in $F$ are merged into background.
    \item $\Psi^{t}_{F+P_4}$: a ternary segmentation network for left kidney and right kidney only based on $F$ and $P_4$. Note that the liver, spleen, pancreas labels in $F$ are merged into background.
    \item $\Psi^{mc}_{All}$: a multi-class segmentation network based on $F$, $P_1$, $P_2$, $P_3$ and $P_4$.
\end{itemize}

\subsection{Training procedure}
For training the above networks except $\Psi^{mc}_{All}$, we use the regular CE loss, regular Dice loss, and their combination. For training the network $\Psi^{mc}_{All}$, when involves partial labels we need to invoke the marginal CE loss, marginal Dice loss, and their combination. Further, for $\Psi^{mc}_{All}$ we experiment the use of exclusion Dice loss and exclusion CE loss. 


Considering the impact of the varying axial resolutions of different data sets in the original CT image on the training process, we resample the 3D CT image to $(1.5\times 1.5\times 3)mm^3$ and then extract the patch with the shape $[190,190,48]$ as input to illustrate the merit of our loss functions. For comparison, we use the same parameter settings in all networks; therefore there is no inference time difference among them. During training, we use 250 batches per epoch and 2 patches per batch. In order to ensure the stability of model training, we set the proportion of patches that contain foreground in each batch to be at least 33\%. The initial learning rate of the network is 1e-1. Whenever the loss reduction is less than 1e-3 in consecutive 10 epochs, the learning rate decays by 20\%. 


We train 3D nnUNet~\citep{isensee2018nnu} for all segmentation networks. We choose the 3D nnUNet because it is known to be a state-of-the-art segmentation network. While there are other network architectures~\citep{tmipld} that might achieve comparable performance, we expect similar empirical observations from our ablation studies even based on the other networks. 

For the network $\Psi^{mc}_{All}$, we train it in two stages in order to prevent the instability caused by large loss value at the beginning of the training. In the first stage, we only use the fully annotated dataset $F$. The goal is to minimize the regular loss function
using the Adam optimizer. The purpose of the first phase is to give the network an initial weight on multi-class segmentation in order to prevent the large loss value when applying the marginal loss functions.
In the second stage, each epoch is trained jointly using the union of five datasets. In each epoch, we randomly select 500 
patches from each training dataset with a batch size of 2. Depending on the source of the slice, we use either the regular loss, if from $F$, or the marginal loss and the exclusion loss, if from $P_i(i\in \{1,2,3,4\})$. In actual experiment, the first stage consists of 120 epochs and the second stage 80 epochs. 


\begin{table*}[t]
    \caption{The Dice coefficients obtained by deep segmentation networks under different loss combinations and on different datasets.}
    \resizebox{\linewidth}{!}{
    \begin{tabular}{l|rrrrrrrrrr|r}
    \hline
        \multicolumn{8}{c}{$\Psi^{mc}_F$: Multiclass ($F$)}\\ \hline
     Loss & Liver$\in F$ & Liver$\in{P_{1}}$& Spleen$\in F$ & Spleen$\in{P_{2}}$& Pancreas$\in{F}$ &Pancreas$\in{P_3}$ &L Kidney$\in{F}$ &R Kidney$\in{F}$ &L Kidney$\in{P_4}$ &R Kidney$\in{P_4}$ & All \\ \hline
rCE     & $.945 \pm .013$ & $.819 \pm .027$ & $.855 \pm .019$ & $.917 \pm .005$ & $.768 \pm .080$ & $.679 \pm .043$ & $.873 \pm .007$ & $.866 \pm .008$ & $.865 \pm .018$ & $.873 \pm .019$ & .846 \\
rDC     & $.945 \pm .014$ & $.837 \pm .031$ & $.857 \pm .018$ & $.914 \pm .007$ & $.768 \pm .012$ & $.673 \pm .047$ & $.720 \pm .005$ & $.821 \pm .007$ & $.812 \pm .016$ & $.917 \pm .012$ & .826 \\
rCE+rDC & $.960 \pm .004$ & $.850 \pm .022$ & $.859 \pm .022$ & $.918 \pm .005$ & $.802 \pm .007$ & $.695 \pm .042$ & $.929 \pm .013$ & $.939 \pm .012$ & $.889 \pm .010$ & $.903 \pm .008$ & .874  \\ \hline \hline
    &\multicolumn{2}{c|}{  $\Psi^{b}_{P_1}$} & \multicolumn{2}{c|}{  $\Psi^{b}_{P_2}$} & \multicolumn{2}{c|}{  $\Psi^{b}_{P_3}$} & \multicolumn{4}{c|}{  $\Psi^{t}_{P_4}$}&\\ \hline
     Loss & Liver$\in F$ & Liver$\in{P_{1}}$& Spleen$\in F$ & Spleen$\in{P_{2}}$& Pancreas$\in{F}$ &Pancreas$\in{P_3}$ &L Kidney$\in{F}$ &R Kidney$\in{F}$ &L Kidney$\in{P_4}$ &R Kidney$\in{P_4}$ & All \\ \hline
rCE     & $.917 \pm .013$ & $.872 \pm .007$ & $.768 \pm .030$ & $.938 \pm .014$ & $.673 \pm .020$ & $.720 \pm .041$ & $.821 \pm .008$ & $.812 \pm .014$ & $.917 \pm .006$ & $.913 \pm .018$ & .835 \\
rDC     & $.931 \pm .027$ & $.883 \pm .008$ & $.817 \pm .027$ & $.940 \pm .015$ & $.670 \pm .019$ & $.715 \pm .041$ & $.817 \pm .009$ & $.807 \pm .012$ & $.908 \pm .007$ & $.900 \pm .017$ & .839 \\
rCE+rDC & $.938 \pm .027$ & $.904 \pm .007$ & $.830 \pm .025$ & $.954 \pm .011$ & $.687 \pm .020$ & $.728 \pm .042$ & $.815 \pm .013$ & $.813 \pm .013$ & $.924 \pm .005$ & $.917 \pm .012$ & .851  \\ \hline \hline
&\multicolumn{2}{c|}{  $\Psi^{b}_{F+P_1}$} & \multicolumn{2}{c|}{  $\Psi^{b}_{F+P_2}$} & \multicolumn{2}{c|}{  $\Psi^{b}_{F+P_3}$} & \multicolumn{4}{c|}{  $\Psi^{t}_{F+P_4}$}&\\ \hline \hline
     Loss & Liver$\in F$ & Liver$\in{P_{1}}$& Spleen$\in F$ & Spleen$\in{P_{2}}$& Pancreas$\in{F}$ &Pancreas$\in{P_3}$ &L Kidney$\in{F}$ &R Kidney$\in{F}$ &L Kidney$\in{P_4}$ &R Kidney$\in{P_4}$ & All \\ \hline
rCE     & $.950 \pm .009$ & $.875 \pm .010$ & $.817 \pm .019$ & $.943 \pm .011$ & $.789 \pm .012$ & $.734 \pm .005$ & $.883 \pm .006$ & $.917 \pm .004$ & $.937 \pm .011$ & $.920 \pm .013$ & .877 \\
rDC     & $.950 \pm .006$ & $.890 \pm .011$ & $.863 \pm .014$ & $.941 \pm .011$ & $.778 \pm .009$ & $.700 \pm .005$ & $.867 \pm .005$ & $.933 \pm .012$ & $.925 \pm .015$ & $.938 \pm .014$ & .879 \\
rCE+rDC & $.960 \pm .012$ & $.899 \pm .008$ & $.869 \pm .014$ & $.945 \pm .011$ & $\underline{.823} \pm .007$ & $.753 \pm .006$ & $.917 \pm .005$ & $.940 \pm .012$ & $.947 \pm .007$ & $.950 \pm .011$ & .900 \\
     
     \hline \hline
    \multicolumn{8}{c}{ $\Psi^{mc}_{All}$: Multiclass ($F+P_1+P_2+P_3+P_4$) }\\ \hline
     Loss & Liver$\in F$ & Liver$\in{P_{1}}$& Spleen$\in F$ & Spleen$\in{P_{2}}$& Pancreas$\in{F}$ &Pancreas$\in{P_3}$ &L Kidney$\in{F}$ &R Kidney$\in{F}$ &L Kidney$\in{P_4}$ &R Kidney$\in{P_4}$ & All \\ \hline
mCE     & $.920 \pm .013$ & $.877 \pm .018$ & $.857 \pm .018$ & $.941 \pm .013$ & $.772 \pm .007$ & $.748 \pm .038$ & $.900 \pm .006$ & $.867 \pm .007$ & $.918 \pm .009$ & $.925 \pm .017$ & .873 \\
mDC     & $.949 \pm .008$ & $.901 \pm .013$ & $.860 \pm .011$  & $.948 \pm .009$ & $.778 \pm .006$ & $.725 \pm .050$ & $.878 \pm .007$ & $.869 \pm .007$ & $.923 \pm .010$ & $.925 \pm .012$ & .876 \\
mCE+mDC & $\underline{.965} \pm .012$ & $\underline{.954} \pm .012$ & $\underline{.891} \pm .015$ & $\underline{.966} \pm .010$ & $.807 \pm .007$ & $\underline{.791} \pm .057$ & $\underline{.942} \pm .012$ & $\underline{.948} \pm .013$ & $\underline{.974} \pm .012$ & $\textbf{.974} \pm .019$ & \underline{.921} \\
mCE+mDC+eCE+eDC      & $\textbf{.969} \pm .012$ & $\textbf{.957} \pm .009$ & $\textbf{.924} \pm .009$ & $\textbf{.970} \pm .008$ & $\textbf{.836} \pm .006$ & $\textbf{.808} \pm .041$ & $\textbf{.946} \pm .012$ & $\textbf{.952} \pm .013$ & $\textbf{.978} \pm .013$ & $\underline{.972} \pm .004$ & \textbf{.931}    \\ \hline
     
    \end{tabular}}
    \label{tab1}
\end{table*}

\begin{table*}[t]
    \caption{The Hausdorff distances obtained by deep segmentation networks under different loss combinations and on different datasets.}
    \resizebox{\linewidth}{!}{
    \begin{tabular}{l|rrrrrrrrrr|r}
    \hline
    
    \multicolumn{8}{c}{  $\Psi^{mc}_F$: Multiclass ($F$)}\\ \hline
     Loss & Liver$\in F$ & Liver$\in{P_{1}}$& Spleen$\in F$ & Spleen$\in {P_{2}}$& Pancreas$\in {F}$ &Pancreas$\in{P_3}$ &L Kidney$\in{F}$ &R Kidney$\in{F}$ &L Kidney$\in{P_4}$ &R Kidney$\in{P_4}$ & All  \\ \hline
rCE     & $\textbf{2.14} \pm 1.69$ & $23.31 \pm 7.25$ & $16.76 \pm 7.00$ & $8.81 \pm 6.12$ & $3.68 \pm 1.46$ & $23.15 \pm 3.92$ & $2.31 \pm 0.59$ & $3.63 \pm 0.35$ & $9.12 \pm 11.58$ & $15.32 \pm 20.84$ & 10.82 \\
rDC     & $2.44 \pm 2.19$ & $23.61 \pm 4.96$ & $19.32 \pm 8.86$ & $8.71 \pm 6.64$ & $3.67 \pm 2.04$ & $23.75 \pm 4.31$ & $2.14 \pm 0.30$ & $3.65 \pm 0.20$ & $8.76 \pm 7.26$ & $7.37 \pm 7.55$  & 10.34 \\
rCE+rDC & $3.21 \pm 1.72$ & $17.36 \pm 3.64$ & $\textbf{16.11} \pm 6.98$ & $8.71 \pm 6.40$ & $6.31 \pm 1.29$ & $21.37 \pm 4.75$ & $2.17 \pm 0.14$ & $3.31 \pm 0.07$ & $8.50 \pm 6.88$ & $6.25 \pm 6.81$  & 9.33\\ \hline \hline
    &\multicolumn{2}{c|}{  $\Psi^{b}_{P_1}$} & \multicolumn{2}{c|}{  $\Psi^{b}_{P_2}$} & \multicolumn{2}{c|}{  $\Psi^{b}_{P_3}$} & \multicolumn{4}{c|}{  $\Psi^{t}_{P_4}$}&\\ \hline
     Loss & Liver$\in F$ & Liver$\in{P_{1}}$& Spleen$\in F$ & Spleen$\in {P_{2}}$& Pancreas$\in {F}$ &Pancreas$\in{P_3}$ &L Kidney$\in{F}$ &R Kidney$\in{F}$ &L Kidney$\in{P_4}$ &R Kidney$\in{P_4}$ & All  \\ \hline
rCE     & $17.32 \pm 3.90$ & $6.31 \pm 3.94$ & $28.32 \pm 9.56$ & $3.76 \pm 0.35$ & $19.36 \pm 3.40$ & $6.55 \pm 4.38$ & $15.38 \pm 5.25$ & $16.47 \pm 5.41$ & $5.07 \pm 7.62$ & $6.32 \pm 21.42$ & 12.49 \\
rDC     & $12.85 \pm 4.37$ & $7.04 \pm 3.44$ & $22.15 \pm 7.00$ & $1.59 \pm 0.47$ & $17.55 \pm 4.54$ & $6.98 \pm 3.05$ & $23.65 \pm 3.52$ & $19.13 \pm 5.26$ & $6.14 \pm 0.31$ & $6.70 \pm 0.37$ & 12.38 \\
rCE+rDC & $18.76 \pm 3.42$ & $\underline{4.00} \pm 3.06$ & $25.67 \pm 7.31$ & $1.13 \pm 0.20$ & $18.36 \pm 4.17$ & $5.46 \pm 3.79$ & $13.66 \pm 5.37$ & $17.33 \pm 7.02$ & $\textbf{1.02} \pm 0.20$ & $1.89 \pm 0.22$ & 10.73  \\ \hline \hline
&\multicolumn{2}{c|}{  $\Psi^{b}_{F+P_1}$} & \multicolumn{2}{c|}{  $\Psi^{b}_{F+P_2}$} & \multicolumn{2}{c|}{  $\Psi^{b}_{F+P_3}$} & \multicolumn{4}{c|}{  $\Psi^{t}_{F+P_4}$}&\\ \hline \hline
     Loss & Liver$\in F$ & Liver$\in{P_{1}}$& Spleen$\in F$ & Spleen$\in {P_{2}}$& Pancreas$\in {F}$ &Pancreas$\in{P_3}$ &L Kidney$\in{F}$ &R Kidney$\in{F}$ &L Kidney$\in{P_4}$ &R Kidney$\in{P_4}$ & All  \\ \hline
rCE     & $6.25 \pm 1.69$ & $8.22 \pm 3.29$  & $30.19 \pm 7.60$ & $2.17 \pm 0.38$ & $13.72 \pm 1.37$ & $9.21 \pm 3.46$  & $7.13 \pm 5.52$ & $8.23 \pm 0.93$  & $7.13 \pm 7.92$ & $6.33 \pm 20.87$ & 9.86 \\
rDC     & $6.49 \pm 1.14$ & $11.25 \pm 3.50$ & $\underline{16.61} \pm 7.27$ & $2.24 \pm 0.20$ & $15.17 \pm 1.17$ & $21.34 \pm 4.42$ & $3.21 \pm 0.34$ & $6.12 \pm 0.63$  & $6.23 \pm 1.14$ & $7.21 \pm 0.63$ & 9.59 \\
rCE+rDC & $\underline{2.63} \pm 0.94$ & $7.49 \pm 3.05$  & $16.85 \pm 7.27$ & $1.65 \pm 0.17$ & $8.16 \pm 0.89$  & $8.56 \pm 3.64$  & $3.46 \pm 0.30$ & $10.70 \pm 2.06$ & $2.24 \pm 0.34$ & $4.66 \pm 0.97$ & 6.64 \\
     \hline \hline
    \multicolumn{8}{c}{ $\Psi^{mc}_{All}$: Multiclass ($F+P_1+P_2+P_3+P_4$) }\\ \hline
     Loss & Liver$\in F$ & Liver$\in{P_{1}}$& Spleen$\in F$ & Spleen$\in {P_{2}}$& Pancreas$\in{F}$ &Pancreas$\in{P_3}$ &L Kidney$\in{F}$ &R Kidney$\in{F}$ &L Kidney$\in{P_4}$ &R Kidney$\in{P_4}$ & All  \\ \hline
mCE     & $8.32 \pm 3.86$ & $15.16 \pm 4.88$ & $17.84 \pm 7.12$ & $2.24 \pm 0.58$ & $12.17 \pm 0.81$ & $8.19 \pm 3.59$ & $4.97 \pm 0.73$ & $15.55 \pm 5.19$ & $6.18 \pm 7.52$ & $7.52 \pm 7.27$ & 9.81 \\
mDC     & $3.72 \pm 3.42$ & $12.71 \pm 3.46$ & $23.62 \pm 6.92$ & $2.44 \pm 0.10$ & $12.36 \pm 0.91$ & $7.18 \pm 3.98$ & $8.19 \pm 0.54$ & $8.85 \pm 6.02$  & $9.16 \pm 7.18$ & $6.55 \pm 7.69$ & 9.48 \\
mCE+mDC & $2.71 \pm 1.16$ & $\textbf{2.94} \pm 2.90$  & $21.67 \pm 7.56$ & $\underline{1.05} \pm 0.09$ & $\underline{4.49} \pm 0.93$  & $\underline{4.92} \pm 3.48$ & $\underline{1.68} \pm 0.29$ & $\underline{1.52} \pm 0.18$  & $\underline{1.77} \pm 0.74$ & $\textbf{1.58} \pm 0.34$ & \underline{4.43} \\
mCE+mDC+eCE+eDC     & $2.84 \pm 1.53$ & $4.04 \pm 2.64$ & $17.58 \pm 7.27$ & $\textbf{1.00} \pm 0.09$ & $\textbf{3.24} \pm 0.69$ & $\textbf{3.96} \pm 3.27$ & $\textbf{1.43} \pm 0.14$  & $\textbf{1.28} \pm 0.07$  & $3.13 \pm 0.58$  & $\underline{1.68} \pm 0.68$ & \textbf{4.02}  \\ \hline
    \end{tabular}}
    \label{hausdorffTab1}
\end{table*}
\subsection{Ablation studies}

We use two standard metrics for gauging the performance of a segmentation method: Dice coefficient and Hausdorff distance (HD). A higher Dice coefficient or a lower HD means a better segmentation result.
Table \ref{tab1} shows the mean and standard deviation of Dice coefficients of the results obtained by the deep segmentation networks under different loss combinations and with different dataset usages, from which we make the following observations.

\textbf{The effect of pooling together more data.} 
The experimental results obtained by the models jointly trained from combinations of the datasets $F$ and $P_i(i \in \{1,2,3,4\})$ are generally better than those by the models trained from a single labeled dataset alone. As shown in Table~\ref{tab1} and Table \ref{hausdorffTab1}, when comparing the performance of $\Psi^{b}_{F+P_i}$ vs $\Psi^{b}_{P_i}(i \in \{1,2,3,4\})$, the former generally outperforms the latter. For example, when using rCE+rDC as the loss, the mean Dice coefficient is boosted from .851 to .900 (the according HD is reduced by 37.5\%). When comparing the performance of $\Psi^{mc}_{F+P_i}(i \in \{1,2,3,4\})$ vs $\Psi^{mc}_{F}$, again the former is better than the latter, the mean dice coefficient is increased from .874 to .900 (the according HD is reduced by 28.7\%). 

\textbf{The importance of CE and Dice losses.} When comparing the importance of CE and Dice losses, in general, it is inconclusive which one is better, depending on the setup. For example, the Dice loss works better on liver segmentation while the CE loss significantly outperforms the Dice loss on left kidney segmentation. Also fusing CE and Dice losses is in general beneficial in terms of our results as it usually brings a gain in segmentation performance. For example, when using $\Psi^{b}_{F}$, the average dice loss reaches .874 for rCE+rDC, while that for rCE and rDC is .846 and .826, respectively.

\begin{table*}[t]
    \caption{The Dice coefficients and Hausdorff distances obtained by the segmentation network $\Psi^{mc}_{All}$ using different loss weight combinations.}
    \resizebox{\linewidth}{!}{
    \begin{tabular}{c|rrrrrrrrrr|r}
    \hline
     mLoss:eLoss & Liver$\in F$ & Liver$\in{P_{1}}$& Spleen$\in F$ & Spleen$\in{P_{2}}$& Pancreas$\in{F}$ &Pancreas$\in{P_3}$ &L Kidney$\in{F}$ &R Kidney$\in{F}$ &L Kidney$\in{P_4}$ &R Kidney$\in{P_4}$ & All  \\ \hline
4:1 & $.962 \pm .011$ & $.931 \pm .029$ & $.884 \pm .019$ & $.954 \pm .013$ & $.775 \pm .007$ & $\underline{.795} \pm .058$ & $.935 \pm .013$ & $.939 \pm .012$ & $.960 \pm .014$ & $.962 \pm .013$ & .910 \\
3:1 & $.964 \pm .007$ & $.952 \pm .017$ & $.890 \pm .015$ & $\underline{.968} \pm .010$ & $.792 \pm .008$ & $.789 \pm .055$ & $.936 \pm .007$ & $.938 \pm .013$ & $.967 \pm .013$ & $.964 \pm .008$ & .916 \\
2:1 & $\textbf{.970} \pm .004$ & $\textbf{.957} \pm .009$ & $.894 \pm .018$ & $\textbf{.970} \pm .007$ & $.833 \pm .005$ & $\textbf{.808} \pm .038$ & $.934 \pm .010$ & $.948 \pm .014$ & $.974 \pm .013$ & $\underline{.969} \pm .013$ & .926 \\
1:1 & $.965 \pm .006$ & $\underline{.954} \pm .015$ & $.893 \pm .016$ & $.966 \pm .009$ & $\textbf{.844} \pm .018$ & $.792 \pm .059$ & $\textbf{.953} \pm .009$ & $\textbf{.959} \pm .004$ & $.977 \pm .020$ & $\textbf{.972} \pm .007$ & $\underline{.928}$ \\
\textbf{1:2} & $\underline{.969} \pm .012$ & $\textbf{.957} \pm .009$ & $\textbf{.924} \pm .009$ & $\textbf{.970} \pm .008$ & $\underline{.836} \pm .006$ & $\textbf{.808} \pm .041$ & $\underline{.946} \pm .012$ & $\underline{.952} \pm .013$ & $\textbf{.978} \pm .013$ & $\textbf{.972} \pm .004$ & \textbf{.931} \\
1:3 & $.968 \pm .009$ & $\underline{.954} \pm .013$ & $\underline{.910} \pm .017$ & $.966 \pm .008$ & $.783 \pm .011$ & $.790 \pm .056$ & $.945 \pm .011$ & $.950 \pm .012$ & $.970 \pm .014$ & $.965 \pm .015$ & .920 \\
1:4 & $.966 \pm .008$ & $.953 \pm .016$ & $.887 \pm .016$ & $.965 \pm .010$ & $.767 \pm .022$ & $.782 \pm .059$ & $.944 \pm .011$ & $.949 \pm .016$ & $.954 \pm .014$ & $.957 \pm .005$ & .913 \\
1:0 & $.965 \pm .012$ & $\underline{.954} \pm .012$ & $.891 \pm .015$ & $.966 \pm .010$ & $.807 \pm .007$ & $.791 \pm .057$ & $.942 \pm .012$ & $.948 \pm .013$ & $.974 \pm .012$ & $.974 \pm .019$ & .921 \\
0:1 & $.967 \pm .012$ & $.930 \pm .035$ & $.904 \pm .020$ & $.958 \pm .011$ & $.785 \pm .015$ & $.678 \pm .057$ & $.926 \pm .008$ & $.934 \pm .006$ & $.950 \pm .019$ & $.941 \pm .018$ & .897  \\ \hline        
\hline
4:1 & $2.89 \pm 0.69$ & $4.39 \pm 1.92$ & $21.43 \pm 7.82$ & $1.41 \pm 0.40$ & $6.76 \pm 2.10$ & $8.42 \pm 3.90$ & $1.85 \pm 0.10$  & $2.01 \pm 0.25$  & $8.12 \pm 8.32$  & $4.39 \pm 2.40$ & 6.17  \\
3:1 & $2.51 \pm 0.40$ & $4.17 \pm 4.42$ & $19.47 \pm 7.79$ & $\textbf{1.00} \pm 0.00$ & $5.92 \pm 2.29$ & $5.11 \pm 3.50$ & $1.90 \pm 0.09$  & $2.01 \pm 0.25$  & $4.18 \pm 1.95$  & $3.75 \pm 0.43$ & 5.00 \\
2:1 & $\underline{1.81} \pm 0.20$ & $4.05 \pm 4.91$ & $22.89 \pm 7.86$ & $\textbf{1.00} \pm 0.00$ & $\underline{3.44} \pm 0.60$ & $\textbf{3.96} \pm 3.18$ & $2.81 \pm 1.80$  & $1.50 \pm 0.12$  & $\textbf{1.25} \pm 0.60$  & $\underline{1.60} \pm 0.95$ & \underline{4.43}  \\
1:1 & $1.98 \pm 0.21$ & $\textbf{2.93} \pm 3.09$ & $21.63 \pm 8.35$ & $\underline{1.05} \pm 0.09$ & $8.72 \pm 3.87$ & $5.17 \pm 3.34$ & $1.58 \pm 0.17$  & $8.04 \pm 5.61$  & $1.79 \pm 1.92$  & $1.66 \pm 0.30$ & 5.46  \\
1:2 & $2.83 \pm 1.53$ & $4.04 \pm 2.64$ & $\underline{17.58} \pm 7.27$ & $\textbf{1.00} \pm 0.09$ & $\textbf{3.24} \pm 0.69$ & $\textbf{3.96} \pm 3.27$ & $\underline{1.43} \pm 0.14$  & $\underline{1.28} \pm 0.07$  & $3.13 \pm 0.08$  & $1.68 \pm 0.68$ & \textbf{4.02}  \\
1:3 & $\textbf{1.41} \pm 0.41$ & $3.03 \pm 2.82$ & $21.50 \pm 9.73$ & $\textbf{1.00} \pm 0.00$ & $8.02 \pm 3.34$ & $5.28 \pm 3.46$ & $\textbf{1.41} \pm 0.14$  & $\textbf{1.00} \pm 0.13$  & $6.76 \pm 0.62$  & $3.13 \pm 0.79$ & 5.25 \\
1:4 & $2.19 \pm 0.51$ & $3.14 \pm 3.17$ & $21.88 \pm 7.94$ & $\underline{1.05} \pm 0.09$ & $8.42 \pm 3.82$ & $5.38 \pm 3.44$ & $12.18 \pm 8.95$ & $1.43 \pm 0.14$  & $8.76 \pm 0.62$  & $4.14 \pm 0.79$ & 6.86  \\
1:0 & $2.71 \pm 1.16$ & $\underline{2.94} \pm 2.90$ & $21.67 \pm 7.56$ & $\underline{1.05} \pm 0.09$ & $4.49 \pm 0.93$ & $\underline{4.92} \pm 3.48$ & $1.68 \pm 0.29$  & $\underline{1.52} \pm 0.18$  & $\underline{1.77} \pm 0.74$  & $\textbf{1.58} \pm 0.34$ & 4.43  \\
0:1 & $2.86 \pm 1.56$ & $6.08 \pm 7.66$ & $\textbf{12.95} \pm 9.59$ & $1.21 \pm 0.05$ & $5.25 \pm 0.70$ & $8.58 \pm 4.34$ & $2.77 \pm 1.55$  & $8.52 \pm 3.75$ & $12.79 \pm 16.34$ & $8.77 \pm 2.21$ & 6.98   \\ \hline
\end{tabular}
}
\label{Dice_weight}
\end{table*}

\begin{table*}[t]
\centering
    \caption{Data sensitivity:  5 sets of experiments with different number of fully labeled and single labeled data.} 
    \begin{tabular}{c|r|rrrrr|r}
    \hline
     full : partial & Total \# of annotated organs & Liver & Spleen& Pancreas&L Kidney&R Kidney& All  \\ \hline
     
24/00  & 120 & $\textbf{.960} \pm .004$ & $\textbf{.859} \pm .022$ & $\textbf{.802} \pm .007$ & $\textbf{.929} \pm .013$ & $\textbf{.939} \pm .012$ & \textbf{.874} \\
19/05  & 100 & $\underline{.938} \pm .012$ & $\underline{.852} \pm .017$ & $\underline{.784}\pm .058$ & $\underline{.879} \pm .015$ & $\underline{.843} \pm .015$ & \underline{.859} \\
14/10  & 80  & $.930 \pm .013$ & $.843 \pm .020$ & $.602 \pm .045$ & $.876 \pm .015$ & $.840 \pm .009$ & .818 \\
09/15  & 60  & $.902 \pm .017$ & $.812 \pm .021$ & $.605 \pm .047$ & $.851 \pm .013$ & $.821 \pm .004$ & .798 \\
04/20  & 40  & $.888 \pm .014$ & $.732 \pm .017$ & $.595 \pm .048$ & $.851 \pm .013$ & $.803 \pm .005$ & .774 \\ \hline \hline

24/00 & 120 & $\textbf{3.21} \pm 1.72$ & $\textbf{16.11} \pm 6.98$ & $\textbf{6.31} \pm 1.29$ & $\textbf{2.17} \pm 0.14$ & $\textbf{3.31} \pm 0.07$ & \textbf{9.33}  \\
19/05 & 100 & $\underline{8.35} \pm 0.62$ & $24.58 \pm 7.53$ & $\underline{8.72} \pm 0.94$  & $\underline{5.72} \pm 0.53$ & $12.66 \pm 5.03$ & \underline{12.00} \\
14/10 & 80  & $8.75 \pm 0.69$ & $26.14 \pm 7.64$ & $23.75 \pm 3.36$ & $8.15 \pm 0.92$ & $\underline{11.75} \pm 6.43$ & 15.71 \\
09/15  & 60  & $9.01 \pm 1.18$ & $\underline{21.18} \pm 7.99$ & $21.97 \pm 3.93$ & $7.32 \pm 0.29$ & $12.39 \pm 7.62$ & 14.37 \\
04/20  & 40  & $8.99 \pm 1.17$ & $27.25 \pm 6.78$ & $23.76 \pm 3.68$ & $7.32 \pm 0.94$ & $13.75 \pm 4.71$ & $16.21$ \\  \hline

\end{tabular}
\label{sensitivity}
\end{table*}

\begin{table*}[h]
    \caption{Segmentation performance comparison in terms of Dice coefficients and Hausdorff distances between our proposed method and state-of-the-art methods.}
    \resizebox{\linewidth}{!}{
    \begin{tabular}{c|rrrrrrrrrr|r}
    \hline
      Methods & Liver$\in{F}$ & Liver$\in{P_{1}}$& Spleen$\in F$ & Spleen$\in{P_{2}}$& Pancreas$\in{F}$ &Pancreas$\in{P_3}$ &L Kidney$\in{F}$ &R Kidney$\in{F}$ &L Kidney$\in{P_4}$ &R Kidney$\in{P_4}$ & All  \\ \hline
      PaNN\citep{zhou2019prior} & $\textbf{.972} \pm .010$ & $\underline{.950} \pm .006$  & $\underline{.915} \pm .008$ & $\underline{.968} \pm .005$ & $\underline{.780} \pm .011$  & $.754 \pm .036$ & $\underline{.901} \pm .006$ & $\underline{.943} \pm .004$ & $.937 \pm .013$ & $.942 \pm .005$ & .906 \\
      PIPO\citep{tmipld} & $.931 \pm .004$ & $.949 \pm .013$ & $.893 \pm .007$ & $.945 \pm .004$ & $.776 \pm .008$ & $.767 \pm .042$ & $.937 \pm .015$ & $.943 \pm .015$ & $ \underline{.959} \pm .004$ & $\underline{.965} \pm .013$ & \underline{.907} \\ 
      our work $\Psi^{mc}_{All}$ & $\underline{.969} \pm .012$ & $\textbf{.957} \pm .009$ & $\textbf{.924} \pm .009$ & $\textbf{.970} \pm .008$ & $\textbf{.836} \pm .006$ & $\textbf{.808} \pm .041$ & $\textbf{.946} \pm .012$ & $\textbf{.952} \pm .013$ & $\textbf{.978} \pm .013$ & $\textbf{.972} \pm .004$ & \textbf{.931} \\
      \hline
      \hline
      PaNN\citep{zhou2019prior} &$\textbf{1.90} \pm 0.95$ & $\underline{4.07} \pm 2.84$ & $21.37 \pm 5.96$ & $\underline{1.05} \pm 0.09$ & $8.64 \pm 1.11$ & $\underline{5.44} \pm 2.54$ & $3.31 \pm 0.58$ & $\underline{1.30} \pm 0.07$ & $\underline{4.20} \pm 0.80$ & $\underline{1.55} \pm 0.14$ & \underline{5.28} \\
      PIPO\citep{tmipld} & $6.40 \pm 0.79$ & $13.87 \pm 6.36$ & $\underline{20.66} \pm 6.12$ & $2.41 \pm 0.35$ & $\underline{6.18} \pm 1.04$ & $5.98 \pm 3.62$ & $\underline{2.32} \pm 0.33$ & $1.31 \pm 0.08$ & $6.79 \pm 1.53$ & $\textbf{1.02} \pm 0.05$  & $6.69$ \\
      our work $\Psi^{mc}_{All}$ & $\underline{2.84} \pm 1.53$ & $\textbf{4.04} \pm 2.64$ & $\textbf{17.58} \pm 7.27$ & $\textbf{1.00} \pm 0.09$ & $\textbf{3.24} \pm 0.69$ & $\textbf{3.96} \pm 3.27$ & $\textbf{1.43} \pm 0.14$  & $\textbf{1.28} \pm 0.07$  & $\textbf{3.13} \pm 0.58$  & $1.68 \pm 0.68$ & \textbf{4.02}  \\
      \hline
    \end{tabular}
    }
\label{otherwork}
\end{table*}

\textbf{The combined effect of data pooling and using marginal loss.}  It is evident that the segmentation network $\Psi^{mc}_{All}$ exhibits a significant performance gain, enabled by joint training on the five datasets. It  brings a 4.7\% increases (.921 vs .874) in average dice coefficient for test images when compared with $\Psi^{mc}_{F}$, which is trained on $F$ alone when using the dice loss and CE. Specifically, it brings an average 5.45\% improvement on liver segmentation (.965 vs .960 on $F$ test images and .954 vs .850 on $P_1$ test images), an average 4.0\% improvement on spleen segmentation (.891 vs .859 on $F$ test images and .966 vs .918 on $P_2$ test images), an average 5.05\% improvement on pancreas segmentation (.807 vs .802 on $F$ test images and .791 vs .695 on $P_3$ test images), and an average 4.45\% improvement on kidney segmentation (.945 vs .934 on $F$ test images and .974 vs .896 on $P_4$ test images).

\textbf{The effect of exclusion loss.} In addition, the exclusion loss brings significant performance boosting. The final results have been effectively improved by an average of 1.0\% increases of Dice coefficient compared to the results obtained without the exclusion loss. This confirms that our proposed exclusion loss can promote the proper learning of the mutual exclusion between two labels. But it should be noted that exclusion loss is more like an auxiliary loss for partial label learning.

In sum, with the help of our newly proposed marginal loss and exclusion loss which enable the joint training of both fully labelled and partially labelled dataset, it brings a 3.1\% increase (.931 vs .900) in dice coefficient. Such a performance improvement is essentially \textbf{a free boost} because these datasets are existent and already labeled.

\textbf{Hausdorff distance.}
Table \ref{hausdorffTab1} shows the mean Hausdorff distance of the testing results, from which similar observations are made. Notably, jointly training from the five datasets, enabled by the marginal loss, can effectively increase the performances, especially it reduces the average distance from 9.33 to 4.43 (a 52.5\% reduction) when using the Dice loss. Adding exclusion dice can further improve the performances (4.43 to 4.02, another 9.3\% reduction). The main reason for the big HD values for say spleen $\in F$ is that sometime a small part of predicted spleen segmentation appears in non-spleen region. This does not affect the Dice coefficient but creates an outlier HD value.

\textbf{The impact of loss weight}. In order to further explore the impact of marginal loss and exclusion loss on the performance, we set up the training of a series of models to understand the influence of the weight ratio of marginal and exclusion losses. All the models are trained on the union of $F$ and all the partially-annotated datasets. We experiment with ten different weight ratios: 4:1, 3:1, 2:1, 1:1, 1:2, 1:3, 1:4, 1:0, and 0:1. The dice coefficients and Hausdorff distances are reported in Table \ref{Dice_weight}. Results demonstrate that a weight ratio of 1:2 achieves the best results on almost all the metrics. It is interesting to observe that, when only using exclusion loss (experiment with a weight of 0:1), there is nearly no performance improvement on pancreas and kidney comparing with $\Psi^{mc}_{F}$, which uses only $F$ for training (as in Tables \ref{tab1} and \ref{hausdorffTab1}). This indicates that exclusion loss is more suitable as an auxiliary loss to be used with marginal loss together.

\textbf{The effect of the number of annotations.} Finally, we perform a group of tests to measure the sensitivity of performance with the number of data annotation increases. We randomly split the fully annotated dataset $F$ into a training set with 24 samples and a testing set with six samples and leave the testing set untouched. In the five sets of experiments reported in Table~\ref{sensitivity} , we alter the training set by replacing some fully labeled data with single labeled data, while keeping the total number of the training data unchanged. For example, for a `14/10' split, we have 14 fully labels images with 5 organs, and the rest of 10 images are further randomly divided into 5 single-label groups of 2 images. For the 1st group, we can use its liver annotation. Similarly we use only the spleen, pancreas, left kidney, and right kidney labels for the 2nd to the 5th groups, respectively. As a result, we have a total of 14*5+2*5=80 annotated organs. Results in Table~\ref{sensitivity} confirm that the dice coefficient consistently decreases as the amount of annotation decreases, which is as expected.

\subsection{Comparison with state-of-the-art}
Our model is also compared with the other partially-supervised segmentation networks. The results are shown in Table~\ref{otherwork}. The Prior-aware Neural Network (PaNN) refers to the work by Zhou et al.~\citep{zhou2019prior} which adds a prior-aware loss to learn partially labeled data. The pyramid input and pyramid output (PIPO) refers to the work by Fang et al.~\citep{tmipld} which develops a multi-scale structure as well as target adaptive loss to enable learning partially labeled data. Our work achieves a significantly better performance than these two methods. The average Dice reaches 0.931 for our model, while that for PaNN and PIPO is 0.906 and 0.907, respectively. Our method also greatly reduce the mean Hausdorff distance by 24.0\% comparing with PaNN and 40.0\% comparing with PIPO. Specifically, our method achieves slight better (except for Liver$\in F$) performance for large organs such as liver and spleen, but it brings a significant performance boost on small organs such as pancreas, left and right kidneys. Our work performs consistently better than the PIPO method on all the organs regardless the datasets, the improvement may be due to the use of 3D model as well as the exclusion loss.

Fig.~\ref{present} presents visualization of sample results of different methods. With the assistance of auxiliary datasets, the performances are significantly improved. Especially, there are situations occurring on all the other methods that the predicted organ region enters a different organ, which results a large HD value. The exclusion loss used in our method can effectively reduce such an error and greatly improve the HD performance. Besides, our method can achieve more meticulous segmentation results on some small organs such as pancreas and kidney, especially when there are small holes around the organ center.

\begin{figure}[htbp]
\centering
\begin{minipage}[t]{0.2\textwidth}
\centering
\includegraphics[width=3.5cm]{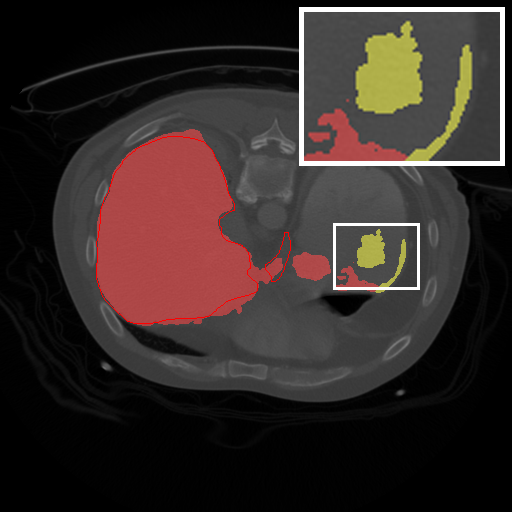}
\end{minipage}
\begin{minipage}[t]{0.2\textwidth}
\centering
\includegraphics[width=3.5cm]{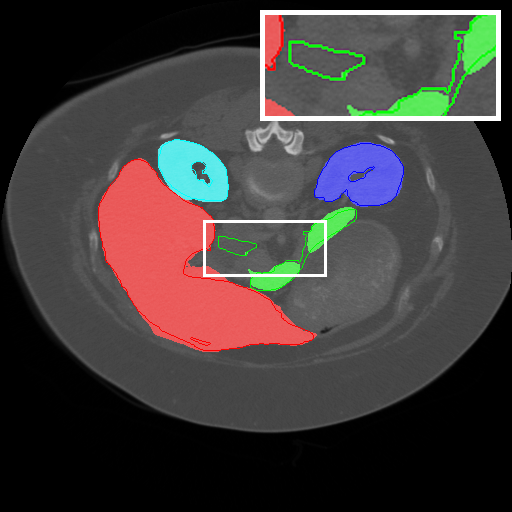}
\end{minipage}
\caption{Failure cases. The figure shows that there are still some regional predictions that have made big mistakes especially in spleen and pancreas.}
\label{badcase}
\end{figure}

\section{Discussions and Conclusions} \label{sec.conc}
In this paper, we propose two new types of loss function that can be used for learning a multi-class segmentation network based on multiple datasets with partial organ labels. The marginal loss enables the learning due to the presence of `merged' labels, while the exclusion loss promotes the learning by adding the mutual exclusiveness as prior knowledge on each labeled image pixel. Our extensive experiments on five benchmark datasets clearly confirm that a significant performance boost is achieved by using marginal loss and exclusion loss. Our method also greatly outperforms existing frameworks for partially annotated data learning.

However, our proposed method is far from perfect. Fig.~\ref{badcase} shows two typical failure cases. In the left image, the background has similar features to liver so the liver prediction on the right side is wrong. In the right image, our method still has some misjudgment on spleen and pancreas. We will generalize the current method for improved segmentation performances by incorporating more knowledge about the organs, such as using shape adversarial prior~\citep{Liverseg}. Furthermore, in future we 
will extend the marginal loss and exclusion loss on other tasks for partially labeled annotated learning and explore the use of other loss functions.



\bibliographystyle{model2-names.bst}\biboptions{authoryear}
\bibliography{refs}

\begin{thebibliography}{53}
\expandafter\ifx\csname natexlab\endcsname\relax\def\natexlab#1{#1}\fi
\providecommand{\url}[1]{\texttt{#1}}
\providecommand{\href}[2]{#2}
\providecommand{\path}[1]{#1}
\providecommand{\DOIprefix}{doi:}
\providecommand{\ArXivprefix}{arXiv:}
\providecommand{\URLprefix}{URL: }
\providecommand{\Pubmedprefix}{pmid:}
\providecommand{\doi}[1]{\href{http://dx.doi.org/#1}{\path{#1}}}
\providecommand{\Pubmed}[1]{\href{pmid:#1}{\path{#1}}}
\providecommand{\bibinfo}[2]{#2}
\ifx\xfnm\relax \def\xfnm[#1]{\unskip,\space#1}\fi
\bibitem[{Berman et~al.(2018)Berman, Rannen~Triki and
  Blaschko}]{berman2018lovasz}
\bibinfo{author}{Berman, M.}, \bibinfo{author}{Rannen~Triki, A.},
  \bibinfo{author}{Blaschko, M.B.}, \bibinfo{year}{2018}.
\newblock \bibinfo{title}{The lov{\'a}sz-softmax loss: A tractable surrogate
  for the optimization of the intersection-over-union measure in neural
  networks}, in: \bibinfo{booktitle}{Proceedings of the IEEE Conference on
  Computer Vision and Pattern Recognition}, pp. \bibinfo{pages}{4413--4421}.
\bibitem[{Binder et~al.(2019)Binder, Tantaoui, Pati, Catena, Set-Aghayan and
  Gabrani}]{binder2019multi}
\bibinfo{author}{Binder, T.}, \bibinfo{author}{Tantaoui, E.M.},
  \bibinfo{author}{Pati, P.}, \bibinfo{author}{Catena, R.},
  \bibinfo{author}{Set-Aghayan, A.}, \bibinfo{author}{Gabrani, M.},
  \bibinfo{year}{2019}.
\newblock \bibinfo{title}{okada2015abdominal}.
\newblock \bibinfo{journal}{Frontiers in Medicine} \bibinfo{volume}{6},
  \bibinfo{pages}{173}.
\bibitem[{Cerrolaza et~al.(2015)Cerrolaza, Reyes, Summers,
  Gonz{\'a}lez-Ballester and Linguraru}]{2015Automatic}
\bibinfo{author}{Cerrolaza, J.J.}, \bibinfo{author}{Reyes, M.},
  \bibinfo{author}{Summers, R.M.}, \bibinfo{author}{Gonz{\'a}lez-Ballester,
  M.{\'A}.}, \bibinfo{author}{Linguraru, M.G.}, \bibinfo{year}{2015}.
\newblock \bibinfo{title}{Automatic multi-resolution shape modeling of
  multi-organ structures}.
\newblock \bibinfo{journal}{Medical image analysis} \bibinfo{volume}{25},
  \bibinfo{pages}{11--21}.
\bibitem[{Chen et~al.(2018)Chen, Dou, Yu, Qin and Heng}]{chen2018voxresnet}
\bibinfo{author}{Chen, H.}, \bibinfo{author}{Dou, Q.}, \bibinfo{author}{Yu,
  L.}, \bibinfo{author}{Qin, J.}, \bibinfo{author}{Heng, P.A.},
  \bibinfo{year}{2018}.
\newblock \bibinfo{title}{Voxresnet: Deep voxelwise residual networks for brain
  segmentation from 3d mr images}.
\newblock \bibinfo{journal}{NeuroImage} \bibinfo{volume}{170},
  \bibinfo{pages}{446--455}.
\bibitem[{Chen et~al.(2012)Chen, Udupa, Bagci, Zhuge and Yao}]{statistical3}
\bibinfo{author}{Chen, X.}, \bibinfo{author}{Udupa, J.K.},
  \bibinfo{author}{Bagci, U.}, \bibinfo{author}{Zhuge, Y.},
  \bibinfo{author}{Yao, J.}, \bibinfo{year}{2012}.
\newblock \bibinfo{title}{Medical image segmentation by combining graph cuts
  and oriented active appearance models}.
\newblock \bibinfo{journal}{IEEE transactions on image processing}
  \bibinfo{volume}{21}, \bibinfo{pages}{2035--2046}.
\bibitem[{Chu et~al.(2013)Chu, Oda, Kitasaka, Misawa, Fujiwara, Hayashi,
  Nimura, Rueckert and Mori}]{chu2013multi}
\bibinfo{author}{Chu, C.}, \bibinfo{author}{Oda, M.},
  \bibinfo{author}{Kitasaka, T.}, \bibinfo{author}{Misawa, K.},
  \bibinfo{author}{Fujiwara, M.}, \bibinfo{author}{Hayashi, Y.},
  \bibinfo{author}{Nimura, Y.}, \bibinfo{author}{Rueckert, D.},
  \bibinfo{author}{Mori, K.}, \bibinfo{year}{2013}.
\newblock \bibinfo{title}{Multi-organ segmentation based on spatially-divided
  probabilistic atlas from 3d abdominal ct images}, in:
  \bibinfo{booktitle}{International conference on medical image computing and
  computer-assisted intervention}, \bibinfo{organization}{Springer}. pp.
  \bibinfo{pages}{165--172}.
\bibitem[{Cootes et~al.(2001)Cootes, Edwards and Taylor}]{statistical2}
\bibinfo{author}{Cootes, T.F.}, \bibinfo{author}{Edwards, G.J.},
  \bibinfo{author}{Taylor, C.J.}, \bibinfo{year}{2001}.
\newblock \bibinfo{title}{Active appearance models}.
\newblock \bibinfo{journal}{IEEE Transactions on pattern analysis and machine
  intelligence} \bibinfo{volume}{23}, \bibinfo{pages}{681--685}.
\bibitem[{Cour et~al.(2011)Cour, Sapp and Taskar}]{JMLR:v12:cour11a}
\bibinfo{author}{Cour, T.}, \bibinfo{author}{Sapp, B.},
  \bibinfo{author}{Taskar, B.}, \bibinfo{year}{2011}.
\newblock \bibinfo{title}{Learning from partial labels}.
\newblock \bibinfo{journal}{Journal of Machine Learning Research}
  \bibinfo{volume}{12}, \bibinfo{pages}{1501--1536}.
\newblock \URLprefix \url{http://jmlr.org/papers/v12/cour11a.html}.
\bibitem[{Dmitriev and Kaufman(2019)}]{Dmitriev_2019_CVPR}
\bibinfo{author}{Dmitriev, K.}, \bibinfo{author}{Kaufman, A.E.},
  \bibinfo{year}{2019}.
\newblock \bibinfo{title}{Learning multi-class segmentations from single-class
  datasets}, in: \bibinfo{booktitle}{The IEEE Conference on Computer Vision and
  Pattern Recognition (CVPR)}.
\bibitem[{{Fang} and {Yan}(2020)}]{tmipld}
\bibinfo{author}{{Fang}, X.}, \bibinfo{author}{{Yan}, P.},
  \bibinfo{year}{2020}.
\newblock \bibinfo{title}{Multi-organ segmentation over partially labeled
  datasets with multi-scale feature abstraction}.
\newblock \bibinfo{journal}{IEEE Transactions on Medical Imaging} ,
  \bibinfo{pages}{1--1}\DOIprefix\doi{10.1109/TMI.2020.3001036}.
\bibitem[{Gibson et~al.(2018)Gibson, Giganti, Hu, Bonmati, Bandula, Gurusamy,
  Davidson, Pereira, Clarkson and Barratt}]{gibson2018automatic}
\bibinfo{author}{Gibson, E.}, \bibinfo{author}{Giganti, F.},
  \bibinfo{author}{Hu, Y.}, \bibinfo{author}{Bonmati, E.},
  \bibinfo{author}{Bandula, S.}, \bibinfo{author}{Gurusamy, K.},
  \bibinfo{author}{Davidson, B.}, \bibinfo{author}{Pereira, S.P.},
  \bibinfo{author}{Clarkson, M.J.}, \bibinfo{author}{Barratt, D.C.},
  \bibinfo{year}{2018}.
\newblock \bibinfo{title}{Automatic multi-organ segmentation on abdominal ct
  with dense v-networks}.
\newblock \bibinfo{journal}{IEEE transactions on medical imaging}
  \bibinfo{volume}{37}, \bibinfo{pages}{1822--1834}.
\bibitem[{Ginneken et~al.(2011)Ginneken, Schaefer-Prokop and
  Prokop}]{Ginneken2011Computer}
\bibinfo{author}{Ginneken, B.V.}, \bibinfo{author}{Schaefer-Prokop, C.M.},
  \bibinfo{author}{Prokop, M.}, \bibinfo{year}{2011}.
\newblock \bibinfo{title}{Computer-aided diagnosis: How to move from the
  laboratory to the clinic}.
\newblock \bibinfo{journal}{Radiology} \bibinfo{volume}{261},
  \bibinfo{pages}{719--732}.
\bibitem[{Hadsell et~al.(2006)Hadsell, Chopra and
  LeCun}]{hadsell2006dimensionality}
\bibinfo{author}{Hadsell, R.}, \bibinfo{author}{Chopra, S.},
  \bibinfo{author}{LeCun, Y.}, \bibinfo{year}{2006}.
\newblock \bibinfo{title}{Dimensionality reduction by learning an invariant
  mapping}, in: \bibinfo{booktitle}{2006 IEEE Computer Society Conference on
  Computer Vision and Pattern Recognition (CVPR'06)},
  \bibinfo{organization}{IEEE}. pp. \bibinfo{pages}{1735--1742}.
\bibitem[{He et~al.(2015)He, Huang and Jia}]{Baochun2015fully}
\bibinfo{author}{He, B.}, \bibinfo{author}{Huang, C.}, \bibinfo{author}{Jia,
  F.}, \bibinfo{year}{2015}.
\newblock \bibinfo{title}{Fully automatic multi-organ segmentation based on
  multi-boost learning and statistical shape model search}.
\newblock \bibinfo{journal}{CEUR Workshop Proceedings} \bibinfo{volume}{1390},
  \bibinfo{pages}{18--21}.
\bibitem[{He et~al.(2019)He, Yang, Gao, Liu and Yin}]{he2019joint}
\bibinfo{author}{He, Z.F.}, \bibinfo{author}{Yang, M.}, \bibinfo{author}{Gao,
  Y.}, \bibinfo{author}{Liu, H.D.}, \bibinfo{author}{Yin, Y.},
  \bibinfo{year}{2019}.
\newblock \bibinfo{title}{Joint multi-label classification and label
  correlations with missing labels and feature selection}.
\newblock \bibinfo{journal}{Knowledge-Based Systems} \bibinfo{volume}{163},
  \bibinfo{pages}{145--158}.
\bibitem[{Heimann and et~al.({2009})}]{Heimann2009Comparison}
\bibinfo{author}{Heimann, T.}, \bibinfo{author}{et~al.},
  \bibinfo{year}{{2009}}.
\newblock \bibinfo{title}{{Comparison and Evaluation of Methods for Liver
  Segmentation From CT Datasets}}.
\newblock \bibinfo{journal}{{IEEE Transactions on Medical Imaging}}
  \bibinfo{volume}{{28}}, \bibinfo{pages}{{1251--1265}}.
\bibitem[{Heimann and Meinzer(2009)}]{heimann2009statistical}
\bibinfo{author}{Heimann, T.}, \bibinfo{author}{Meinzer, H.P.},
  \bibinfo{year}{2009}.
\newblock \bibinfo{title}{Statistical shape models for 3d medical image
  segmentation: a review}.
\newblock \bibinfo{journal}{Medical image analysis} \bibinfo{volume}{13},
  \bibinfo{pages}{543--563}.
\bibitem[{Heller et~al.(2019)Heller, Sathianathen, Kalapara, Walczak, Moore,
  Kaluzniak, Rosenberg, Blake, Rengel, Oestreich et~al.}]{KiTS}
\bibinfo{author}{Heller, N.}, \bibinfo{author}{Sathianathen, N.},
  \bibinfo{author}{Kalapara, A.}, \bibinfo{author}{Walczak, E.},
  \bibinfo{author}{Moore, K.}, \bibinfo{author}{Kaluzniak, H.},
  \bibinfo{author}{Rosenberg, J.}, \bibinfo{author}{Blake, P.},
  \bibinfo{author}{Rengel, Z.}, \bibinfo{author}{Oestreich, M.}, et~al.,
  \bibinfo{year}{2019}.
\newblock \bibinfo{title}{The kits19 challenge data: 300 kidney tumor cases
  with clinical context, ct semantic segmentations, and surgical outcomes}.
\newblock \bibinfo{journal}{arXiv preprint arXiv:1904.00445} .
\bibitem[{Isensee et~al.(2018)Isensee, Petersen, Klein, Zimmerer, Jaeger, Kohl,
  Wasserthal, Koehler, Norajitra, Wirkert et~al.}]{isensee2018nnu}
\bibinfo{author}{Isensee, F.}, \bibinfo{author}{Petersen, J.},
  \bibinfo{author}{Klein, A.}, \bibinfo{author}{Zimmerer, D.},
  \bibinfo{author}{Jaeger, P.F.}, \bibinfo{author}{Kohl, S.},
  \bibinfo{author}{Wasserthal, J.}, \bibinfo{author}{Koehler, G.},
  \bibinfo{author}{Norajitra, T.}, \bibinfo{author}{Wirkert, S.}, et~al.,
  \bibinfo{year}{2018}.
\newblock \bibinfo{title}{nnu-net: Self-adapting framework for u-net-based
  medical image segmentation}.
\newblock \bibinfo{journal}{arXiv preprint arXiv:1809.10486} .
\bibitem[{Kohlberger et~al.(2011)Kohlberger, Sofka, Zhang, Birkbeck, Wetzl,
  Kaftan, Declerck and Zhou}]{multi_levelset}
\bibinfo{author}{Kohlberger, T.}, \bibinfo{author}{Sofka, M.},
  \bibinfo{author}{Zhang, J.}, \bibinfo{author}{Birkbeck, N.},
  \bibinfo{author}{Wetzl, J.}, \bibinfo{author}{Kaftan, J.},
  \bibinfo{author}{Declerck, J.}, \bibinfo{author}{Zhou, S.K.},
  \bibinfo{year}{2011}.
\newblock \bibinfo{title}{Automatic multi-organ segmentation using
  learning-based segmentation and level set optimization}, in:
  \bibinfo{editor}{Fichtinger, G.}, \bibinfo{editor}{Martel, A.},
  \bibinfo{editor}{Peters, T.} (Eds.), \bibinfo{booktitle}{Medical Image
  Computing and Computer-Assisted Intervention}, \bibinfo{publisher}{Springer
  Berlin Heidelberg}, \bibinfo{address}{Berlin, Heidelberg}. pp.
  \bibinfo{pages}{338--345}.
\bibitem[{Landman et~al.(2017)Landman, Xu, Igelsias, Styner, Langerak and
  Klein}]{landman2017multi}
\bibinfo{author}{Landman, B.}, \bibinfo{author}{Xu, Z.},
  \bibinfo{author}{Igelsias, J.}, \bibinfo{author}{Styner, M.},
  \bibinfo{author}{Langerak, T.}, \bibinfo{author}{Klein, A.},
  \bibinfo{year}{2017}.
\newblock \bibinfo{title}{Multi-atlas labeling beyond the cranial
  vault-workshop and challenge}.
\bibitem[{Lay et~al.(2013)Lay, Birkbeck, Zhang and Zhou}]{Contextmodel}
\bibinfo{author}{Lay, N.}, \bibinfo{author}{Birkbeck, N.},
  \bibinfo{author}{Zhang, J.}, \bibinfo{author}{Zhou, S.K.},
  \bibinfo{year}{2013}.
\newblock \bibinfo{title}{Rapid multi-organ segmentation using context
  integration and discriminative models}, in: \bibinfo{editor}{Gee, J.C.},
  \bibinfo{editor}{Joshi, S.}, \bibinfo{editor}{Pohl, K.M.},
  \bibinfo{editor}{Wells, W.M.}, \bibinfo{editor}{Z{\"o}llei, L.} (Eds.),
  \bibinfo{booktitle}{Information Processing in Medical Imaging},
  \bibinfo{publisher}{Springer Berlin Heidelberg}, \bibinfo{address}{Berlin,
  Heidelberg}. pp. \bibinfo{pages}{450--462}.
\bibitem[{Li et~al.(2018)Li, Gao, Ou and Sun}]{li2018angular}
\bibinfo{author}{Li, Y.}, \bibinfo{author}{Gao, F.}, \bibinfo{author}{Ou, Z.},
  \bibinfo{author}{Sun, J.}, \bibinfo{year}{2018}.
\newblock \bibinfo{title}{Angular softmax loss for end-to-end speaker
  verification}, in: \bibinfo{booktitle}{2018 11th International Symposium on
  Chinese Spoken Language Processing (ISCSLP)}, \bibinfo{organization}{IEEE}.
  pp. \bibinfo{pages}{190--194}.
\bibitem[{Lin et~al.(2017)Lin, Goyal, Girshick, He and
  Doll{\'a}r}]{lin2017focal}
\bibinfo{author}{Lin, T.Y.}, \bibinfo{author}{Goyal, P.},
  \bibinfo{author}{Girshick, R.}, \bibinfo{author}{He, K.},
  \bibinfo{author}{Doll{\'a}r, P.}, \bibinfo{year}{2017}.
\newblock \bibinfo{title}{Focal loss for dense object detection}, in:
  \bibinfo{booktitle}{Proceedings of the IEEE international conference on
  computer vision}, pp. \bibinfo{pages}{2980--2988}.
\bibitem[{Liu et~al.(2016)Liu, Wen, Yu and Yang}]{liu2016large}
\bibinfo{author}{Liu, W.}, \bibinfo{author}{Wen, Y.}, \bibinfo{author}{Yu, Z.},
  \bibinfo{author}{Yang, M.}, \bibinfo{year}{2016}.
\newblock \bibinfo{title}{Large-margin softmax loss for convolutional neural
  networks}, in: \bibinfo{booktitle}{International Conference on Machine
  Learning}, p.~\bibinfo{pages}{7}.
\bibitem[{Liu et~al.(2020)Liu, Gargesha, Qutaish, Zhou, Scott, Yousefi, Lu and
  Wilson}]{liu2020deep}
\bibinfo{author}{Liu, Y.}, \bibinfo{author}{Gargesha, M.},
  \bibinfo{author}{Qutaish, M.}, \bibinfo{author}{Zhou, Z.},
  \bibinfo{author}{Scott, B.}, \bibinfo{author}{Yousefi, H.},
  \bibinfo{author}{Lu, Z.}, \bibinfo{author}{Wilson, D.L.},
  \bibinfo{year}{2020}.
\newblock \bibinfo{title}{Deep learning based multi-organ segmentation and
  metastases segmentation in whole mouse body and the cryo-imaging cancer
  imaging and therapy analysis platform (citap)}, in:
  \bibinfo{booktitle}{Medical Imaging 2020: Biomedical Applications in
  Molecular, Structural, and Functional Imaging},
  \bibinfo{organization}{International Society for Optics and Photonics}. p.
  \bibinfo{pages}{113170V}.
\bibitem[{Lombaert et~al.(2014)Lombaert, Zikic, Criminisi and
  Ayache}]{Lombaert2014Laplacian}
\bibinfo{author}{Lombaert, H.}, \bibinfo{author}{Zikic, D.},
  \bibinfo{author}{Criminisi, A.}, \bibinfo{author}{Ayache, N.},
  \bibinfo{year}{2014}.
\newblock \bibinfo{title}{Laplacian forests: Semantic image segmentation by
  guided bagging}, in: \bibinfo{booktitle}{International Conference on Medical
  Image Computing and Computer-assisted Intervention}.
\bibitem[{Long et~al.(2015)Long, Shelhamer and Darrell}]{Long_2015_CVPR}
\bibinfo{author}{Long, J.}, \bibinfo{author}{Shelhamer, E.},
  \bibinfo{author}{Darrell, T.}, \bibinfo{year}{2015}.
\newblock \bibinfo{title}{Fully convolutional networks for semantic
  segmentation}, in: \bibinfo{booktitle}{The IEEE Conference on Computer Vision
  and Pattern Recognition (CVPR)}.
\bibitem[{Lu et~al.(2012)Lu, Zheng, Birkbeck, Zhang, Kohlberger, Tietjen,
  Boettger, Duncan and Zhou}]{other1}
\bibinfo{author}{Lu, C.}, \bibinfo{author}{Zheng, Y.},
  \bibinfo{author}{Birkbeck, N.}, \bibinfo{author}{Zhang, J.},
  \bibinfo{author}{Kohlberger, T.}, \bibinfo{author}{Tietjen, C.},
  \bibinfo{author}{Boettger, T.}, \bibinfo{author}{Duncan, J.S.},
  \bibinfo{author}{Zhou, S.K.}, \bibinfo{year}{2012}.
\newblock \bibinfo{title}{Precise segmentation of multiple organs in ct volumes
  using learning-based approach and information theory}, in:
  \bibinfo{booktitle}{International Conference on Medical Image Computing and
  Computer-Assisted Intervention}, \bibinfo{organization}{Springer}. pp.
  \bibinfo{pages}{462--469}.
\bibitem[{Okada et~al.(2015)Okada, Linguraru, Hori, Summers, Tomiyama and
  Sato}]{okada2015abdominal}
\bibinfo{author}{Okada, T.}, \bibinfo{author}{Linguraru, M.G.},
  \bibinfo{author}{Hori, M.}, \bibinfo{author}{Summers, R.M.},
  \bibinfo{author}{Tomiyama, N.}, \bibinfo{author}{Sato, Y.},
  \bibinfo{year}{2015}.
\newblock \bibinfo{title}{Abdominal multi-organ segmentation from ct images
  using conditional shape--location and unsupervised intensity priors}.
\newblock \bibinfo{journal}{Medical image analysis} \bibinfo{volume}{26},
  \bibinfo{pages}{1--18}.
\bibitem[{Okada et~al.(2012)Okada, Linguraru, Hori, Suzuki, Summers, Tomiyama
  and Sato}]{okada2012multi}
\bibinfo{author}{Okada, T.}, \bibinfo{author}{Linguraru, M.G.},
  \bibinfo{author}{Hori, M.}, \bibinfo{author}{Suzuki, Y.},
  \bibinfo{author}{Summers, R.M.}, \bibinfo{author}{Tomiyama, N.},
  \bibinfo{author}{Sato, Y.}, \bibinfo{year}{2012}.
\newblock \bibinfo{title}{Multi-organ segmentation in abdominal ct images}, in:
  \bibinfo{booktitle}{2012 Annual International Conference of the IEEE
  Engineering in Medicine and Biology Society}, \bibinfo{organization}{IEEE}.
  pp. \bibinfo{pages}{3986--3989}.
\bibitem[{Ronneberger et~al.(2015)Ronneberger, Fischer and
  Brox}]{ronneberger2015u}
\bibinfo{author}{Ronneberger, O.}, \bibinfo{author}{Fischer, P.},
  \bibinfo{author}{Brox, T.}, \bibinfo{year}{2015}.
\newblock \bibinfo{title}{U-net: Convolutional networks for biomedical image
  segmentation}, in: \bibinfo{booktitle}{International Conference on Medical
  image computing and computer-assisted intervention},
  \bibinfo{organization}{Springer}. pp. \bibinfo{pages}{234--241}.
\bibitem[{Salehi et~al.(2017)Salehi, Erdogmus and
  Gholipour}]{salehi2017tversky}
\bibinfo{author}{Salehi, S.S.M.}, \bibinfo{author}{Erdogmus, D.},
  \bibinfo{author}{Gholipour, A.}, \bibinfo{year}{2017}.
\newblock \bibinfo{title}{Tversky loss function for image segmentation using 3d
  fully convolutional deep networks}, in: \bibinfo{booktitle}{International
  Workshop on Machine Learning in Medical Imaging},
  \bibinfo{organization}{Springer}. pp. \bibinfo{pages}{379--387}.
\bibitem[{Saxena et~al.(2016)Saxena, Sharma, Sharma, Singh and
  Verma}]{Saxena2016Automated}
\bibinfo{author}{Saxena, S.}, \bibinfo{author}{Sharma, N.},
  \bibinfo{author}{Sharma, S.}, \bibinfo{author}{Singh, S.},
  \bibinfo{author}{Verma, A.}, \bibinfo{year}{2016}.
\newblock \bibinfo{title}{An automated system for atlas based multiple organ
  segmentation of abdominal ct images}.
\newblock \bibinfo{journal}{British Journal of Mathematics and Computer
  Science} \bibinfo{volume}{12}, \bibinfo{pages}{1--14}.
\newblock \DOIprefix\doi{10.9734/BJMCS/2016/20812}.
\bibitem[{Schroff et~al.(2015)Schroff, Kalenichenko and
  Philbin}]{schroff2015facenet}
\bibinfo{author}{Schroff, F.}, \bibinfo{author}{Kalenichenko, D.},
  \bibinfo{author}{Philbin, J.}, \bibinfo{year}{2015}.
\newblock \bibinfo{title}{Facenet: A unified embedding for face recognition and
  clustering}, in: \bibinfo{booktitle}{Proceedings of the IEEE conference on
  computer vision and pattern recognition}, pp. \bibinfo{pages}{815--823}.
\bibitem[{Shimizu et~al.(2007)Shimizu, Ohno, Ikegami, Kobatake, Nawano and
  Smutek}]{shimizu2007segmentation}
\bibinfo{author}{Shimizu, A.}, \bibinfo{author}{Ohno, R.},
  \bibinfo{author}{Ikegami, T.}, \bibinfo{author}{Kobatake, H.},
  \bibinfo{author}{Nawano, S.}, \bibinfo{author}{Smutek, D.},
  \bibinfo{year}{2007}.
\newblock \bibinfo{title}{Segmentation of multiple organs in non-contrast 3d
  abdominal ct images}.
\newblock \bibinfo{journal}{International journal of computer assisted
  radiology and surgery} \bibinfo{volume}{2}, \bibinfo{pages}{135--142}.
\bibitem[{Simpson et~al.(2019)Simpson, Antonelli, Bakas, Bilello, Farahani,
  Van~Ginneken, Kopp-Schneider, Landman, Litjens, Menze
  et~al.}]{simpson2019large}
\bibinfo{author}{Simpson, A.L.}, \bibinfo{author}{Antonelli, M.},
  \bibinfo{author}{Bakas, S.}, \bibinfo{author}{Bilello, M.},
  \bibinfo{author}{Farahani, K.}, \bibinfo{author}{Van~Ginneken, B.},
  \bibinfo{author}{Kopp-Schneider, A.}, \bibinfo{author}{Landman, B.A.},
  \bibinfo{author}{Litjens, G.}, \bibinfo{author}{Menze, B.}, et~al.,
  \bibinfo{year}{2019}.
\newblock \bibinfo{title}{A large annotated medical image dataset for the
  development and evaluation of segmentation algorithms}.
\newblock \bibinfo{journal}{arXiv preprint arXiv:1902.09063} .
\bibitem[{Suzuki et~al.(2012)Suzuki, Linguraru and Okada}]{suzuki2012multi}
\bibinfo{author}{Suzuki, M.}, \bibinfo{author}{Linguraru, M.G.},
  \bibinfo{author}{Okada, K.}, \bibinfo{year}{2012}.
\newblock \bibinfo{title}{Multi-organ segmentation with missing organs in
  abdominal ct images}, in: \bibinfo{booktitle}{International Conference on
  Medical Image Computing and Computer-Assisted Intervention},
  \bibinfo{organization}{Springer}. pp. \bibinfo{pages}{418--425}.
\bibitem[{Sykes(2014)}]{Sykes2014Reflections}
\bibinfo{author}{Sykes, J.}, \bibinfo{year}{2014}.
\newblock \bibinfo{title}{Reflections on the current status of commercial
  automated segmentation systems in clinical practice}.
\newblock \bibinfo{journal}{Journal of Medical Radiation Sciences}
  \bibinfo{volume}{61}, \bibinfo{pages}{131–134}.
\bibitem[{Taghanaki et~al.(2019)Taghanaki, Zheng, Zhou, Georgescu, Sharma, Xu,
  Comaniciu and Hamarneh}]{taghanaki2019combo}
\bibinfo{author}{Taghanaki, S.A.}, \bibinfo{author}{Zheng, Y.},
  \bibinfo{author}{Zhou, S.K.}, \bibinfo{author}{Georgescu, B.},
  \bibinfo{author}{Sharma, P.}, \bibinfo{author}{Xu, D.},
  \bibinfo{author}{Comaniciu, D.}, \bibinfo{author}{Hamarneh, G.},
  \bibinfo{year}{2019}.
\newblock \bibinfo{title}{Combo loss: Handling input and output imbalance in
  multi-organ segmentation}.
\newblock \bibinfo{journal}{Computerized Medical Imaging and Graphics}
  \bibinfo{volume}{75}, \bibinfo{pages}{24--33}.
\bibitem[{Tong et~al.(2015)Tong, Wolz, Wang, Gao, Misawa, Fujiwara, Mori,
  Hajnal and Rueckert}]{tong2015discriminative}
\bibinfo{author}{Tong, T.}, \bibinfo{author}{Wolz, R.}, \bibinfo{author}{Wang,
  Z.}, \bibinfo{author}{Gao, Q.}, \bibinfo{author}{Misawa, K.},
  \bibinfo{author}{Fujiwara, M.}, \bibinfo{author}{Mori, K.},
  \bibinfo{author}{Hajnal, J.V.}, \bibinfo{author}{Rueckert, D.},
  \bibinfo{year}{2015}.
\newblock \bibinfo{title}{Discriminative dictionary learning for abdominal
  multi-organ segmentation}.
\newblock \bibinfo{journal}{Medical image analysis} \bibinfo{volume}{23},
  \bibinfo{pages}{92--104}.
\bibitem[{Wang et~al.(2018)Wang, Wang, Zhou, Ji, Gong, Zhou, Li and
  Liu}]{wang2018cosface}
\bibinfo{author}{Wang, H.}, \bibinfo{author}{Wang, Y.}, \bibinfo{author}{Zhou,
  Z.}, \bibinfo{author}{Ji, X.}, \bibinfo{author}{Gong, D.},
  \bibinfo{author}{Zhou, J.}, \bibinfo{author}{Li, Z.}, \bibinfo{author}{Liu,
  W.}, \bibinfo{year}{2018}.
\newblock \bibinfo{title}{Cosface: Large margin cosine loss for deep face
  recognition}, in: \bibinfo{booktitle}{Proceedings of the IEEE Conference on
  Computer Vision and Pattern Recognition}, pp. \bibinfo{pages}{5265--5274}.
\bibitem[{Wang et~al.(2019)Wang, Zhou, Shen, Park, Fishman and
  Yuille}]{wang2019abdominal}
\bibinfo{author}{Wang, Y.}, \bibinfo{author}{Zhou, Y.}, \bibinfo{author}{Shen,
  W.}, \bibinfo{author}{Park, S.}, \bibinfo{author}{Fishman, E.K.},
  \bibinfo{author}{Yuille, A.L.}, \bibinfo{year}{2019}.
\newblock \bibinfo{title}{Abdominal multi-organ segmentation with
  organ-attention networks and statistical fusion}.
\newblock \bibinfo{journal}{Medical image analysis} \bibinfo{volume}{55},
  \bibinfo{pages}{88--102}.
\bibitem[{Wen et~al.(2016)Wen, Zhang, Li and Qiao}]{wen2016discriminative}
\bibinfo{author}{Wen, Y.}, \bibinfo{author}{Zhang, K.}, \bibinfo{author}{Li,
  Z.}, \bibinfo{author}{Qiao, Y.}, \bibinfo{year}{2016}.
\newblock \bibinfo{title}{A discriminative feature learning approach for deep
  face recognition}, in: \bibinfo{booktitle}{European conference on computer
  vision}, \bibinfo{organization}{Springer}. pp. \bibinfo{pages}{499--515}.
\bibitem[{Wolz et~al.(2013)Wolz, Chu, Misawa, Fujiwara, Mori and
  Rueckert}]{altas1}
\bibinfo{author}{Wolz, R.}, \bibinfo{author}{Chu, C.}, \bibinfo{author}{Misawa,
  K.}, \bibinfo{author}{Fujiwara, M.}, \bibinfo{author}{Mori, K.},
  \bibinfo{author}{Rueckert, D.}, \bibinfo{year}{2013}.
\newblock \bibinfo{title}{Automated abdominal multi-organ segmentation with
  subject-specific atlas generation}.
\newblock \bibinfo{journal}{IEEE transactions on medical imaging}
  \bibinfo{volume}{32}, \bibinfo{pages}{1723--1730}.
\bibitem[{Wu et~al.(2015)Wu, Lyu, Hu and Ji}]{wu2015multi}
\bibinfo{author}{Wu, B.}, \bibinfo{author}{Lyu, S.}, \bibinfo{author}{Hu,
  B.G.}, \bibinfo{author}{Ji, Q.}, \bibinfo{year}{2015}.
\newblock \bibinfo{title}{Multi-label learning with missing labels for image
  annotation and facial action unit recognition}.
\newblock \bibinfo{journal}{Pattern Recognition} \bibinfo{volume}{48},
  \bibinfo{pages}{2279--2289}.
\bibitem[{Xiao et~al.(2019)Xiao, Zhu, Liu, Luo, Liu and Zhao}]{Li2019}
\bibinfo{author}{Xiao, L.}, \bibinfo{author}{Zhu, C.}, \bibinfo{author}{Liu,
  J.}, \bibinfo{author}{Luo, C.}, \bibinfo{author}{Liu, P.},
  \bibinfo{author}{Zhao, Y.}, \bibinfo{year}{2019}.
\newblock \bibinfo{title}{Learning from suspected target: Bootstrapping
  performance for breast cancer detection in mammography}, in:
  \bibinfo{booktitle}{Medical Image Computing and Computer Assisted
  Intervention}, \bibinfo{publisher}{Springer International Publishing},
  \bibinfo{address}{Cham}. pp. \bibinfo{pages}{468--476}.
\bibitem[{Xu et~al.(2015)Xu, Burke, Lee, Baucom, Poulose, Abramson and
  Landman}]{xu2015efficient}
\bibinfo{author}{Xu, Z.}, \bibinfo{author}{Burke, R.P.}, \bibinfo{author}{Lee,
  C.P.}, \bibinfo{author}{Baucom, R.B.}, \bibinfo{author}{Poulose, B.K.},
  \bibinfo{author}{Abramson, R.G.}, \bibinfo{author}{Landman, B.A.},
  \bibinfo{year}{2015}.
\newblock \bibinfo{title}{Efficient multi-atlas abdominal segmentation on
  clinically acquired ct with simple context learning}.
\newblock \bibinfo{journal}{Medical image analysis} \bibinfo{volume}{24},
  \bibinfo{pages}{18--27}.
\bibitem[{Yang et~al.(2017)Yang, Xu, Zhou, Georgescu, Chen, Grbic, Metaxas and
  Comaniciu}]{Liverseg}
\bibinfo{author}{Yang, D.}, \bibinfo{author}{Xu, D.}, \bibinfo{author}{Zhou,
  S.K.}, \bibinfo{author}{Georgescu, B.}, \bibinfo{author}{Chen, M.},
  \bibinfo{author}{Grbic, S.}, \bibinfo{author}{Metaxas, D.},
  \bibinfo{author}{Comaniciu, D.}, \bibinfo{year}{2017}.
\newblock \bibinfo{title}{Automatic liver segmentation using an adversarial
  image-to-image network}, in: \bibinfo{editor}{Descoteaux, M.},
  \bibinfo{editor}{Maier-Hein, L.}, \bibinfo{editor}{Franz, A.},
  \bibinfo{editor}{Jannin, P.}, \bibinfo{editor}{Collins, D.L.},
  \bibinfo{editor}{Duchesne, S.} (Eds.), \bibinfo{booktitle}{Medical Image
  Computing and Computer Assisted Intervention}, \bibinfo{publisher}{Springer
  International Publishing}, \bibinfo{address}{Cham}. pp.
  \bibinfo{pages}{507--515}.
\bibitem[{Yu et~al.(2014)Yu, Jain, Kar and Dhillon}]{yu2014large}
\bibinfo{author}{Yu, H.F.}, \bibinfo{author}{Jain, P.}, \bibinfo{author}{Kar,
  P.}, \bibinfo{author}{Dhillon, I.}, \bibinfo{year}{2014}.
\newblock \bibinfo{title}{Large-scale multi-label learning with missing
  labels}, in: \bibinfo{booktitle}{International conference on machine
  learning}, pp. \bibinfo{pages}{593--601}.
\bibitem[{Zhao et~al.(2019)Zhao, Balakrishnan, Durand, Guttag and
  Dalca}]{zhao2019data}
\bibinfo{author}{Zhao, A.}, \bibinfo{author}{Balakrishnan, G.},
  \bibinfo{author}{Durand, F.}, \bibinfo{author}{Guttag, J.V.},
  \bibinfo{author}{Dalca, A.V.}, \bibinfo{year}{2019}.
\newblock \bibinfo{title}{Data augmentation using learned transformations for
  one-shot medical image segmentation}, in: \bibinfo{booktitle}{Proceedings of
  the IEEE conference on computer vision and pattern recognition}, pp.
  \bibinfo{pages}{8543--8553}.
\bibitem[{Zhou et~al.(2019)Zhou, Li, Bai, Wang, Chen, Han, Fishman and
  Yuille}]{zhou2019prior}
\bibinfo{author}{Zhou, Y.}, \bibinfo{author}{Li, Z.}, \bibinfo{author}{Bai,
  S.}, \bibinfo{author}{Wang, C.}, \bibinfo{author}{Chen, X.},
  \bibinfo{author}{Han, M.}, \bibinfo{author}{Fishman, E.},
  \bibinfo{author}{Yuille, A.L.}, \bibinfo{year}{2019}.
\newblock \bibinfo{title}{{Prior-aware neural network for partially-supervised
  multi-organ segmentation}}, in: \bibinfo{booktitle}{Proceedings of the IEEE
  International Conference on Computer Vision}, pp.
  \bibinfo{pages}{10672--10681}.
\bibitem[{Zhu et~al.(2018)Zhu, Xu, Hu, Zhang and Zhao}]{zhu2018multi}
\bibinfo{author}{Zhu, P.}, \bibinfo{author}{Xu, Q.}, \bibinfo{author}{Hu, Q.},
  \bibinfo{author}{Zhang, C.}, \bibinfo{author}{Zhao, H.},
  \bibinfo{year}{2018}.
\newblock \bibinfo{title}{Multi-label feature selection with missing labels}.
\newblock \bibinfo{journal}{Pattern Recognition} \bibinfo{volume}{74},
  \bibinfo{pages}{488--502}.

\end{thebibliography}



\end{document}